\documentclass[runningheads]{llncs}
\usepackage{graphicx}
\usepackage{comment}
\usepackage{amsmath,amssymb} %
\usepackage{color}

\usepackage{xspace}
\usepackage{bbm}
\usepackage{makecell}
\usepackage{subfigure}

\usepackage[breaklinks=true,bookmarks=false]{hyperref}
\def\method{{SOLO}\xspace}
\def\OurMethod{{SOLO}\xspace}

\newcommand{\myparagraph}[1]{{ \noindent \bf #1}}

\newcommand{\ie}{\textit{i}.\textit{e}.}
\newcommand{\eg}{\textit{e}.\textit{g}.}

\begin{document}
\pagestyle{headings}
\mainmatter
\def\ECCVSubNumber{3082}  %

\title{\bf
\Large
        \method:
        Segmenting Objects by Locations\thanks{Correspondence should be addressed to C. Shen.}}
\titlerunning{ECCV-20 submission ID \ECCVSubNumber}
\authorrunning{ECCV-20 submission ID \ECCVSubNumber}
\author{Anonymous ECCV submission}
\institute{Paper ID \ECCVSubNumber}
\titlerunning{SOLO (Accepted to Proc.\ Eur.\ Conf.\ Computer Vision (ECCV), 2020)}
\author{Xinlong Wang\inst{1} %
\and
Tao Kong\inst{2} %
\and
Chunhua Shen\inst{1} %
\and
Yuning Jiang\inst{2}
\and
Lei Li\inst{2}
}
\authorrunning{Wang et al.}
\institute{
The University of Adelaide, Australia
\and
ByteDance AI Lab
}
\maketitle

\begin{abstract}

We present a new, embarrassingly simple approach to instance segmentation. %
Compared to many other dense prediction tasks, \textit{e.g.}, semantic segmentation,
it is the arbitrary number of instances that have made
instance segmentation much more challenging.
In order to predict a mask for each instance, mainstream approaches either
follow the ``detect-then-segment''  strategy (\textit{e.g.}, Mask R-CNN),
or predict
embedding vectors
first then use clustering techniques to group pixels into individual instances.
We view the task of instance segmentation from
a completely new perspective by introducing
the notion of ``instance categories'',
which assigns categories to each pixel within an instance according to the instance's location and size,
thus nicely converting instance segmentation into a
single-shot
classification-solvable problem.
We demonstrate a much simpler and flexible instance segmentation framework
with strong performance,
achieving  \textit{on par} accuracy with Mask R-CNN and
outperforming
recent single-shot instance segmenters in accuracy.
We hope that this
simple and strong framework can serve as a baseline
for many instance-level recognition tasks besides instance segmentation.
Code is available at  \url{https://git.io/AdelaiDet}
\end{abstract}

\section{Introduction}
Instance segmentation is challenging because it requires the correct separation of all objects in an image while also semantically segmenting each instance
at the pixel level.
Objects in an image belong to
a fixed set of
semantic categories, but the number of instances varies.
As a result, semantic segmentation can be
easily
formulated as a dense per-pixel classification problem, while it is
challenging
to predict instance labels directly
following the same paradigm.

To overcome this obstacle, recent instance segmentation methods can be categorized into two groups, \ie, top-down and bottom-up paradigms.
The former approach, namely `detect-then-segment', first detects
bounding boxes and then segments the instance mask
in each bounding box.
The latter approach learns an affinity relation,
assigning an embedding vector to each pixel, by pushing away pixels belonging to different instances and pulling close pixels in the same instance.
A grouping post-processing is then needed to separate instances.
Both these two paradigms are step-wise and {\it indirect}, which either heavily rely on
accurate bounding box detection %
or depend on per-pixel embedding learning and the grouping processing.

\iffalse
\begin{figure}[tbp]
\centering
\subfigure[Mask R-CNN]{
\includegraphics[width=0.35\textwidth]{figures/maskrcnn_framework.pdf}
\label{fig:framework1}
}
\subfigure[\OurMethod]{
\includegraphics[width=0.27\textwidth]{figures/our_framework.pdf}
\label{fig:framework2}
}
\caption{Comparison of the pipelines of Mask R-CNN and the proposed \OurMethod.}
\label{fig:framework_comparison}
\end{figure}
\fi

\begin{figure}[tbp]
\centering
\includegraphics[width=0.81\textwidth]{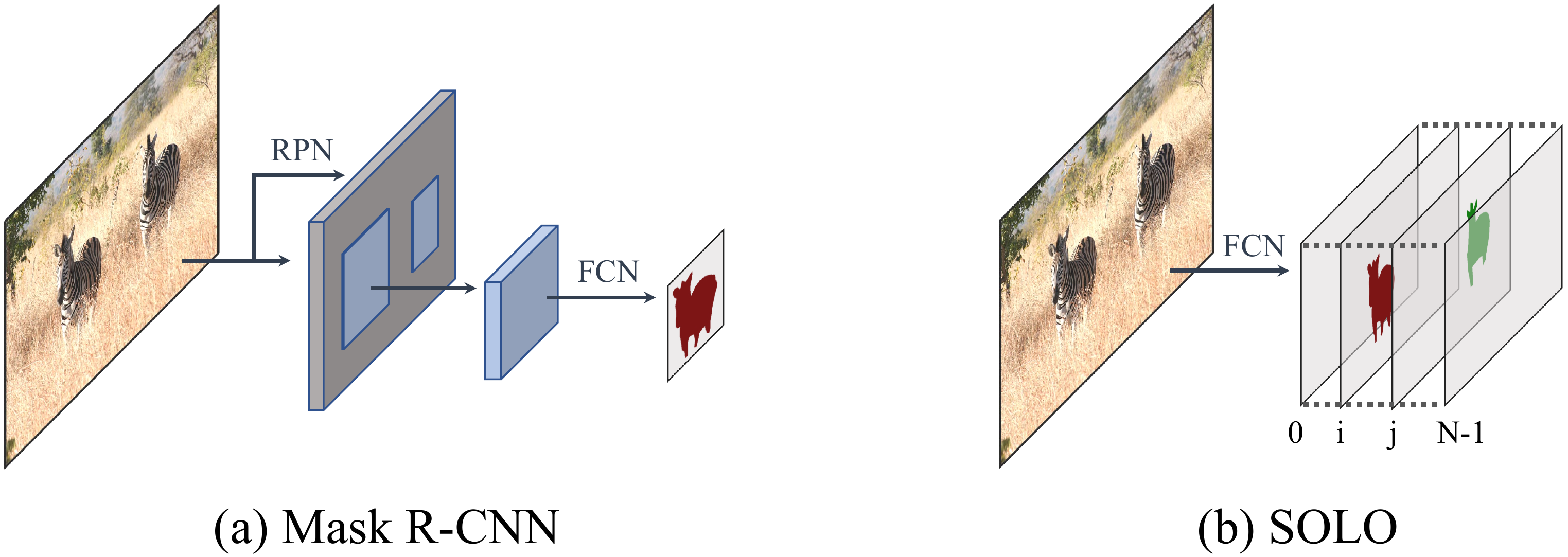}
\caption{Comparison of the pipelines of Mask R-CNN and the proposed \OurMethod.}
\label{fig:framework_comparison}
\end{figure}

In contrast, we aim to directly segment instance masks, under the supervision of full instance mask annotations instead of masks in boxes or
additional
pixel
pairwise
relations. We start by rethinking a question:
\textit{What are the fundamental differences between object instances in an image?}
Take the challenging MS COCO dataset~\cite{coco} for example.
There are in total
$36,780$ objects in the validation subset, $98.3\%$ of object pairs have center distance greater than $30$ pixels.
As for the rest $1.7\%$ of object pairs, $40.5\%$ of them have size ratio greater than 1.5$\times$.
To conclude, in most cases two instances in an image either have different center locations or have different object sizes.
This observation makes one wonder whether we could directly distinguish instances by the center locations and object sizes?

In the closely related field, semantic segmentation,  now
the dominate paradigm leverages a fully convolutional network (FCN) to output dense predictions with $N$ channels. Each output channel is responsible for one of the semantic categories (including background).
Semantic segmentation aims to distinguish different semantic categories.
Analogously,
in this work, we propose to distinguish object instances
in the image
by introducing
the notion of ``\textit{instance categories}'', \ie, the quantized center locations and object sizes, which enables to \textbf{s}egment \textbf{o}bjects by \textbf{lo}cations, thus the name of our method,
{\bf
\OurMethod.
}

\textbf{Locations}
An image can be divided into a grid of $S \times S$ cells, thus leading to $S^2$ center location classes. According to the coordinates of the object center, an object instance is assigned to one of the grid cells, as its center location category.
Note that grids are used conceptually to assign location category for each pixel.
Each output channel is responsible for one of the center location categories, and the corresponding channel map should predict the instance mask of the object belonging to that location.
Thus,
structural geometric information is naturally preserved in the spatial matrix with dimensions of height by width.
Unlike DeepMask~\cite{deepmask} and TensorMask~\cite{Chen_2019_ICCV}, which run in a dense sliding-window manner and segment an object in a fixed local patch, our method
naturally outputs accurate masks for all scales of instances without the limitation of (anchor) box locations and scales.

In essence, an instance location category approximates
the location of the object center of an instance.
Thus, by classification of each pixel into its instance location category, it is equivalent to predict the object center of each pixel in the latent space.
The importance here of converting
the location prediction task into classification
is that, with
classification it is much
more straightforward and easier to model
varying number of instances using a fixed number of channels, at the same time not relying on post-processing like grouping or learning embeddings.

\textbf{Sizes}
To distinguish instances with different object sizes,
we employ the feature pyramid network
(FPN)~\cite{fpn}, so as to
assign objects %
of
different sizes to different levels of feature maps.
Thus, all the object instances are separated regularly, enabling to classify objects by ``instance categories''.
Note that FPN was designed for the purposes of detecting objects of different sizes
in an image.

In the sequel, we empirically show that FPN is one of the core components for our method and has a profound impact on the segmentation performance, especially objects of varying sizes being presented.

With the  proposed \OurMethod framework, we are able to optimize the network
in an end-to-end
fashion
for the instance segmentation task
using %
mask
annotations solely, and perform pixel-level instance segmentation out of the restrictions of local box detection and pixel grouping.
For the first time, we demonstrate a very simple instance segmentation approach achieving {\it on par} results to the dominant ``detect-then-segment'' method
on the challenging COCO dataset~\cite{coco} with diverse scenes and
semantic classes.
Additionally,
we showcase the generality of our framework via the task of instance contour detection, by viewing the instance edge contours as a one-hot binary mask, with almost no
modification \OurMethod can generate reasonable instance contours.
The proposed \OurMethod  only needs to solve two pixel-level classification tasks,
thus it may be possible to borrow some of the recent advances in semantic segmentation
for improving \OurMethod.
\textit{The embarrassing simplicity and strong performance of the proposed
\OurMethod  method
may predict its application to a wide range of instance-level recognition
tasks}.

\subsection{Related Work}
 We review some instance segmentation works that are closest to ours.

\myparagraph{Top-down Instance Segmentation.}
The methods that segment object instance in a priori bounding box fall into the typical top-down paradigm.
FCIS~\cite{fcis} assembles the position-sensitive score maps within the  region-of-interests (ROIs) generated by a region proposal network (RPN) to predict instance masks.
Mask R-CNN~\cite{maskrcnn} extends the Faster R-CNN detector~\cite{fasterrcnn} by adding a branch for segmenting the object instances within the detected bounding boxes.
Based on Mask R-CNN, PANet~\cite{panet} further enhances the feature representation to improve the accuracy,
Mask Scoring R-CNN~\cite{maskscoringrcnn} adds a mask-IoU branch to predict the quality of the predicted mask and scoring the masks to improve the performance.
HTC~\cite{chen2019hybrid} interweaves box and mask branches for a joint multi-stage processing.
TensorMask~\cite{Chen_2019_ICCV} adopts the dense sliding window paradigm to segment the instance in the local window for each pixel with a predefined number of windows  and scales.
In contrast to the top-down methods above, our \OurMethod is totally box-free thus not being restricted by (anchor) box locations and scales, and naturally benefits from the inherent advantages of FCNs.

\myparagraph{Bottom-up Instance Segmentation.}
This category  of the approaches generate instance masks by grouping the pixels into
an
arbitrary number of object instances presented in an image.
In~\cite{associativeembedding}, pixels are grouped into instances using the learned associative embedding.
A discriminative loss function~\cite{de2017semantic} learns pixel-level instance embedding efficiently, by pushing away pixels belonging to different instances and pulling close pixels in the same instance.
SGN~\cite{SGN17} decomposes the instance segmentation problem into a sequence of sub-grouping problems.
SSAP~\cite{Gao_2019_ICCV} learns a pixel-pair affinity pyramid, the probability that two pixels belong to the same instance, and sequentially generates instances by a cascaded graph partition.
Typically
 bottom-up methods lag behind in accuracy compared to top-down methods, especially on the dataset with diverse scenes.
Instead of exploiting pixel pairwise relations
\OurMethod directly learns %
with
the instance mask annotations solely during training,
and predicts instance masks
end-to-end without
grouping post-processing.

\myparagraph{Direct Instance Segmentation.}
To
our knowledge, no prior methods directly train with
mask
annotations solely,  and predict instance masks and semantic categories in one shot without the need of grouping post-processing.
Several recently proposed methods may be viewed as the `semi-direct' paradigm.
AdaptIS~\cite{adaptis} first predicts point proposals, and then sequentially generates the mask for the object located at the detected point proposal.
PolarMask~\cite{polarmask} proposes to use the polar representation
to encode masks
and transforms per-pixel mask prediction to distance regression.
They both do not need bounding boxes for training but are either being step-wise or founded on compromise, \eg, coarse parametric representation of masks.
Our \OurMethod takes an image as input, directly outputs instance masks and corresponding class probabilities, in a fully convolutional, box-free and grouping-free paradigm.

\section{Our Method:  \OurMethod}

\begin{figure*}[t]
\centering
    \includegraphics[width=0.995\linewidth]{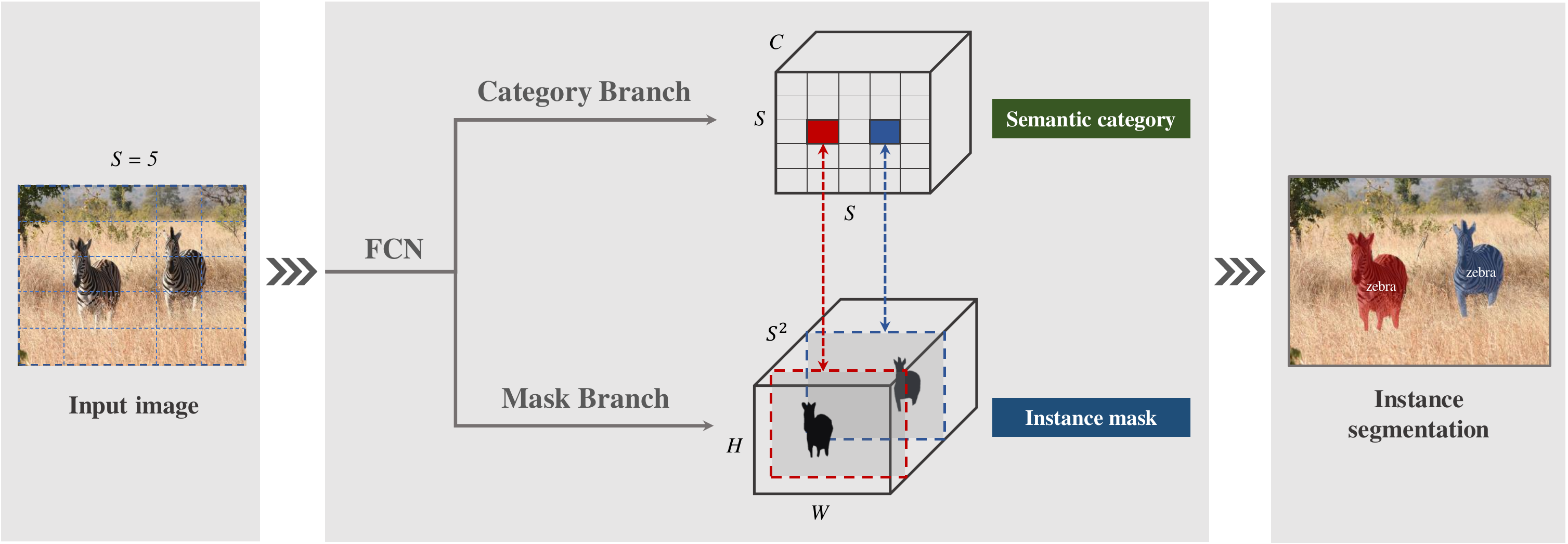}
   \caption{\textbf{\OurMethod framework.}
    We reformulate the instance segmentation as two sub-tasks:
    category prediction and instance mask generation problems.
    An input image is divided into a uniform grids, \textit{i.e.}, $S$$\times$$S$.
    Here we illustrate the grid with $S = 5$.
    If the center of an object falls into a grid cell, that grid cell
    is responsible for predicting the semantic category (top)
    and masks of instances (bottom).
    We do not show the
    feature pyramid network (FPN) here for simpler
    illustration.
   }
\label{fig:pipline}
\end{figure*}

\subsection{Problem Formulation}
The central idea of \OurMethod framework is to reformulate the instance segmentation as two simultaneous category-aware prediction
problems. Concretely, our system divides the input image into a uniform grids, \textit{i.e.}, $S$$\times$$S$. If the center of an object falls into a grid cell, that grid cell
is responsible for 1) predicting the semantic category as well as 2) segmenting that object instance.

\subsubsection{Semantic Category}
For each grid, our \method predicts the $C$-dimensional output to indicate the semantic class probabilities,  where $C$ is the number of classes. These probabilities are conditioned on the grid cell. If we divide the input image into $S$$\times$$S$ grids, the output space will be  $S$$\times$$S$$\times$$C$, as shown in Figure~\ref{fig:pipline} (top). This design is based on the assumption that each cell of the $S$$\times$$S$ grid must belong to one individual instance, thus only belonging to one semantic category. During inference, the $C$-dimensional output indicates the class probability for each object instance.

\subsubsection{Instance Mask}
In parallel with the semantic category prediction, each positive grid cell will also generate the corresponding instance mask. For an input image $I$, if we divide it into $S$$\times$$S$ grids, there will be at most $S^2$ predicted masks in total. We explicitly encode these masks at the third dimension (channel) of a
3D
 output  tensor. Specifically, the instance mask output will have $H_I$$\times$$W_I$$\times$$S^2$ dimension. The $k^{th}$ channel will be responsible to segment instance at grid ($i$, $j$), where $k=i\cdot S + j$ (with $i$ and $j$ zero-based)\footnote{We also show an equivalent and more efficient implementation in Section \ref{sec:decouple}.}. To this end,  a one-to-one correspondence is established between the semantic category and class-agnostic mask (Figure~\ref{fig:pipline}).

A direct approach to predict the instance mask is to adopt the fully convolutional networks, like FCNs in semantic segmentation~\cite{fcn}. However the conventional convolutional operations are \textit{spatially  invariant} to some degree.
Spatial invariance is desirable for some tasks such as image classification
as it introduces robustness. However,
here
we need a model that is \textit{spatially variant},
or in more precise words, position sensitive,
since our segmentation masks are conditioned on the grid cells and must be separated by different feature channels.

Our solution is very simple:
at the beginning of the network, we directly feed normalized pixel coordinates to the
networks, inspired  by  `CoordConv' operator~\cite{coordconv}. Specifically, we create a tensor of same spatial size as input that contains pixel coordinates,
which are  normalized to $[-1, 1]$.
This tensor is then concatenated to the input features and passed to the following layers. By simply %
giving  the convolution access to its own input coordinates, we add the spatial functionality to the conventional FCN model.
It should be noted that CoordConv is not the only choice. For example the semi-convolutional operators~\cite{semiconv}  may  be competent, but we employ CoordConv for its simplicity and being easy to implement. If the original feature tensor is of size $H$$\times$$W$$\times$$D$, the size of new tensor becomes $H$$\times$$W$$\times$$(D+2)$, in which the last two channels are $x$-$y$ pixel coordinates. For more information on CoordConv, we refer readers to~\cite{coordconv}.

\myparagraph{Forming Instance Segmentation.}
In \method, the category prediction and the corresponding mask are naturally associated by their reference grid cell, \ie, $k=i\cdot S + j$. Based on this, we can directly form the final instance segmentation result for each grid. The raw instance segmentation results are generated by gathering all grid results. Finally, non-maximum-suppression (NMS) is used to
obtain the final instance segmentation results. No other post processing operations are
needed.

\subsection{Network Architecture}

\begin{figure}[t!]
\begin{center}
    \includegraphics[width=0.6\linewidth]{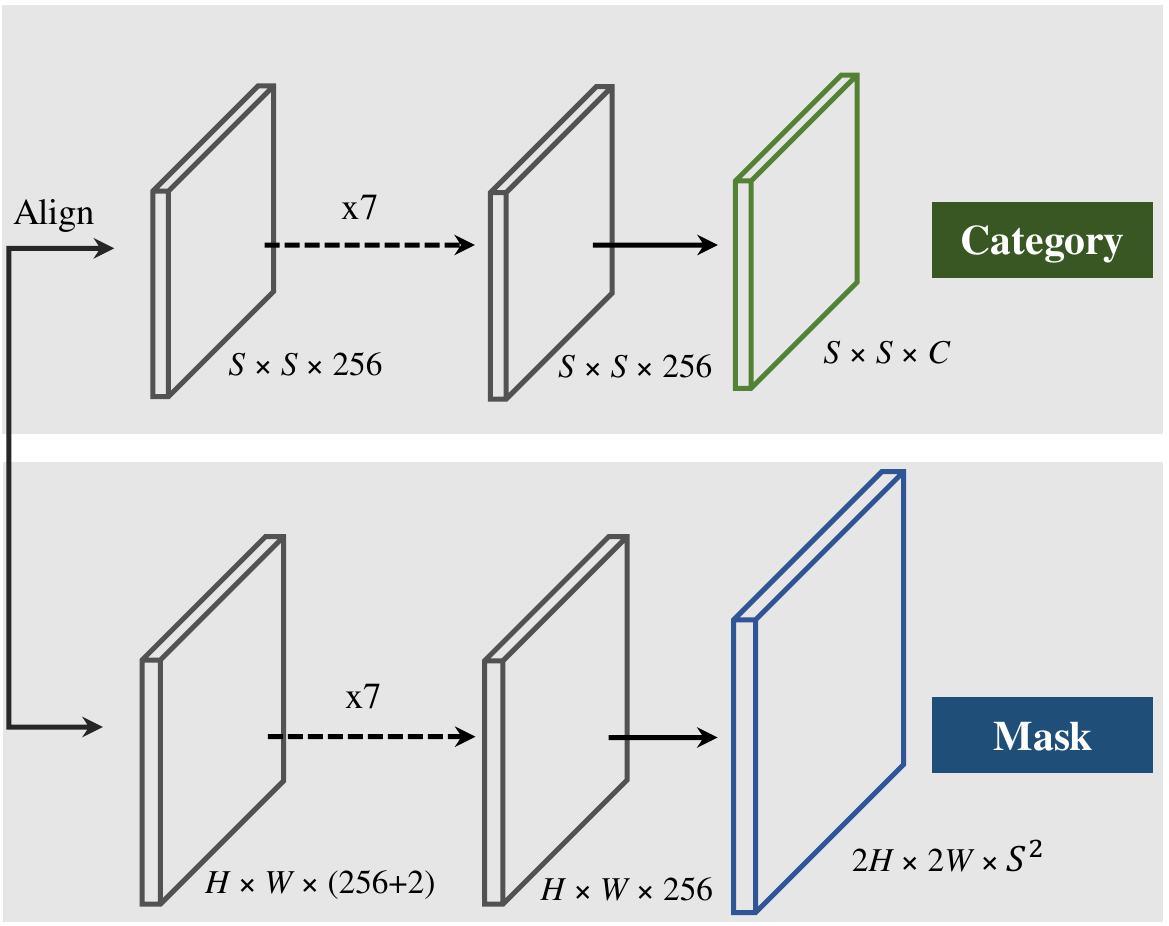}
\end{center}
   \caption{\textbf{\method Head architecture}. At each FPN feature level, we attach two sibling sub-networks, one for instance category prediction (top) and one for instance mask segmentation (bottom).
   In the mask branch, we concatenate the $x$, $y$ coordinates and the original features to
   encode spatial information.
   Here numbers denote spatial resolution and channels.
   In the figure, we assume 256 channels as an example.
   Arrows denote either convolution or interpolation.
   `Align' means
   bilinear interpolation.
   During inference, the mask branch outputs
   are
   further upsampled to the original image size.}
\label{fig:head}
\end{figure}

\OurMethod attaches to a convolutional backbone.
We use FPN ~\cite{fpn}, which generates a pyramid of feature maps with different sizes with  a  fixed number of channels (usually 256-d) for each level. These maps are used as input for each prediction head: semantic category and instance mask. Weights for the head are shared across different levels. Grid number may varies at different pyramids.  Only the last %
conv is not shared in this scenario.

To demonstrate the generality and effectiveness of our approach, we instantiate \method with multiple architectures. The differences include: (a) the \textit{backbone} architecture used for feature extraction, (b) the network \textit{head} for computing the instance segmentation results, and (c) training \textit{loss function} used to optimize the model. Most of the experiments are based on the \textit{head} architecture as shown in Figure~\ref{fig:head}. We also utilize different variants to further study the generality. We note that our instance segmentation heads have a straightforward structure. More complex designs have the potential to improve performance but are not the focus of this work.

\subsection{\OurMethod Learning}

\subsubsection{Label Assignment}
For category prediction branch, the network needs to give the object category probability for each of $S$$\times$$S$ grid. Specifically, grid $(i, j)$ is considered as a positive sample if it falls into the \textit{center region} of any ground truth mask, Otherwise it is a negative sample. Center sampling is effective in recent works of object detection~\cite{fcos,foveabox}, and here we also utilize a similar technique for mask category classification. Given the mass center $(c_x, c_y)$, width $w$, and height $h$ of the ground truth mask, the center region is controlled by constant scale factors $\epsilon$: $(c_x, c_y, \epsilon w, \epsilon h)$. We set $\epsilon=0.2$ and there are on average 3 positive samples for each ground truth mask.

Besides the label for instance category, we also have a binary segmentation mask for each positive sample. Since there are $S^2$ grids, we also have $S^2$ output masks for each image. For each positive samples, the corresponding target binary mask will be annotated. One may be concerned that the order of masks will impact the mask prediction branch, however, we show that the most simple row-major order works well for our method.

\subsubsection{Loss Function}
We define our training loss function as follows:
\begin{equation}
\label{eq:loss_all}
\begin{aligned}
L = L_{cate} + \lambda L_{mask},
\end{aligned}
\end{equation}
where $L_{cate}$ is the conventional Focal Loss~\cite{focalloss} for semantic category classification. $L_{mask}$ is the loss for mask prediction:
\begin{equation}
\label{eq:loss}
\begin{aligned}
L_{mask}  = \frac{1}{N_{pos}} \sum_{k}  \mathbbm{1}_{\{\mathbf{p}^{*}_{ i,j  } > 0\}} d_{mask} (\mathbf{m}_{k}, \mathbf{m}^{*}_{k}),
\end{aligned}
\end{equation}
Here indices $ i =  \lfloor k/S\rfloor, j = k \, {\rm mod} \, S $,
if we index the grid cells  (instance category labels) from left to right and top to down.
$N_{pos}$ denotes the number of positive samples, $\mathbf{p}^{*}$ and $\mathbf{m}^{*}$ represent category and mask target
respectively.
$\mathbbm{1}$ is the indicator function, being 1 if $\mathbf{p}^{*}_{i, j} > 0$ and 0 otherwise.

We have compared different implementations of $d_{mask}(\cdot,\cdot)$: Binary Cross Entropy (BCE), Focal Loss~\cite{focalloss} and Dice Loss~\cite{vnet}. Finally, we employ Dice Loss for its effectiveness and stability in training.
$\lambda$ in Equation~\eqref{eq:loss_all} is set to 3.
The Dice Loss is defined as
\begin{equation}
\begin{aligned}
L_{Dice} = 1 - D(\mathbf{p}, \mathbf{q}),
\end{aligned}
\end{equation}
where $D$ is the dice coefficient which is defined as
\begin{equation}
\begin{aligned}
D(\mathbf{p}, \mathbf{q}) = \frac{2\sum_{x,y}(\mathbf{p}_{x,y} \cdot  \mathbf{q}_{x,y})}{\sum_{x,y}\mathbf{p}^2_{x,y} + \sum_{x,y}\mathbf{q}^2_{x,y}}.
\end{aligned}
\end{equation}
Here $\mathbf{p}_{x,y}$ and $\mathbf{q}_{x,y}$ refer to the value of pixel located at $(x, y)$ in
predicted soft mask $\mathbf{p}$ and ground truth mask $\mathbf{q}$.

\subsection{Inference}

The inference of \method is very straightforward. Given an input image, we forward it through the backbone network and FPN, and obtain the category score $\mathbf{p}_{i,j}$ at grid $(i, j)$ and the corresponding masks $\mathbf{m}_k$, where $k=i \cdot  S + j$.
We first use a confidence threshold of $0.1$ to filter out predictions with low confidence. Then we select the top $500$ scoring masks and feed them into the NMS operation. We
use a threshold of $0.5$ to convert predicted soft masks to binary masks. %

\myparagraph{Maskness.} We calculate maskness for each predicted mask, which represents the quality and confidence of mask prediction $\rm{maskness} = \frac{1}{N_{f}} \sum_{i}^{N_{f}} {\mathbf{p}_{i}}$.
Here $N_f$ the number of foreground pixels of the predicted soft mask $\mathbf{p}$, \ie, the pixels that have values greater than threshold $0.5$.
The classification score for each prediction is multiplied by the maskness as the final confidence score.

\section{Experiments}

We present experimental results on the MS COCO instance segmentation benchmark~\cite{coco}, and report ablation studies by evaluating on the 5k \texttt{val2017} split. For our main results, we report COCO mask AP on the
\texttt{test}-\texttt{dev} split, which has no public labels and is evaluated on the evaluation server.

\begin{table*}[bt]
\centering
\caption{\textbf{Instance segmentation mask AP} (\%)  on the COCO \texttt{test}-\texttt{dev}. All entries are \textit{single-model} results. Here we adopt the ``6$\times$'' schedule (72 epochs), following~\cite{Chen_2019_ICCV}.
Mask R-CNN$ ^*$ is our improved version with scale augmentation and longer training time. D-\OurMethod means Decoupled SOLO as introduced in Section \ref{sec:decouple}.
}
\begin{tabular}{ l | l |cccccc}
& backbone &AP & AP$_{50}$ & AP$_{75}$&AP$_{S}$ & AP$_{M}$ & AP$_{L}$\\
\Xhline{1pt}
\emph{two-stage:} &&&&&&&\\
~~~~MNC~\cite{mnc}                    &Res-101-C4  &24.6   &44.3   &24.8   &4.7    &25.9   &43.6 \\
~~~~FCIS~\cite{fcis}                  &Res-101-C5  &29.2   &49.5   & $-$      &7.1    &31.3   &50.0 \\
~~~~Mask R-CNN~\cite{maskrcnn}        &Res-101-FPN &35.7  &58.0   &37.8   &15.5   &38.1   &52.4 \\
~~~~MaskLab+~\cite{masklab}           &Res-101-C4  &37.3  & 59.8   & 39.6   & 16.9   & 39.9   &53.5 \\
~~~~Mask R-CNN$^*$                    &Res-101-FPN &37.8  &59.8   &40.7   &20.5   &40.4   &49.3 \\
\hline
\emph{one-stage:} &&&&&&&\\
~~~~TensorMask~\cite{Chen_2019_ICCV}  &Res-50-FPN & 35.4  & 57.2   & 37.3  &16.3   &36.8   &49.3 \\
~~~~TensorMask~\cite{Chen_2019_ICCV}  &Res-101-FPN &37.1  &59.3   &39.4   &17.4   &39.1   &51.6 \\
~~~~YOLACT~\cite{yolact}              &Res-101-FPN &31.2  &50.6   &32.8   &12.1   &33.3   &47.1 \\
~~~~PolarMask~\cite{polarmask}        &Res-101-FPN &30.4  &51.9   &31.0   &13.4   &32.4   &42.8 \\
\hline\hline
\emph{ours:} &&&&&&&\\
~~~~\textbf{\OurMethod}                            &Res-50-FPN  &  36.8 & 58.6 & 39.0 & 15.9 & 39.5 & 52.1\\
~~~~\textbf{\OurMethod}                            &Res-101-FPN  &  37.8 & 59.5 & 40.4 & 16.4 & 40.6 & 54.2\\
~~~~\textbf{D-\OurMethod ~}                            &Res-101-FPN  &  38.4 & 59.6 & 41.1 & 16.8 & 41.5 & 54.6\\
~~~~\textbf{D-\OurMethod ~}                            &Res-DCN-101-FPN &  40.5 & 62.4  & 43.7  & 17.7  & 43.6  & 59.3  \\
\end{tabular}

\label{tab:sota}
\end{table*}

\myparagraph{Training details.}
\method is trained with stochastic gradient descent (SGD). We use synchronized SGD over 8 GPUs with a total of 16 images per mini-batch.
Unless otherwise specified, all models are trained for  36 epochs with an initial learning rate of $0.01$, which is then
divided by 10 at 27th and again at 33th epoch. Weight decay of $0.0001$ and momentum of $ 0.9$ are used. All models are initialized from ImageNet
pre-trained weights. We use scale jitter where the shorter
image side is randomly sampled from 640 to 800 pixels, following~\cite{Chen_2019_ICCV}.

\begin{figure*}[h]
\begin{center}
    \includegraphics[width=0.98\linewidth]{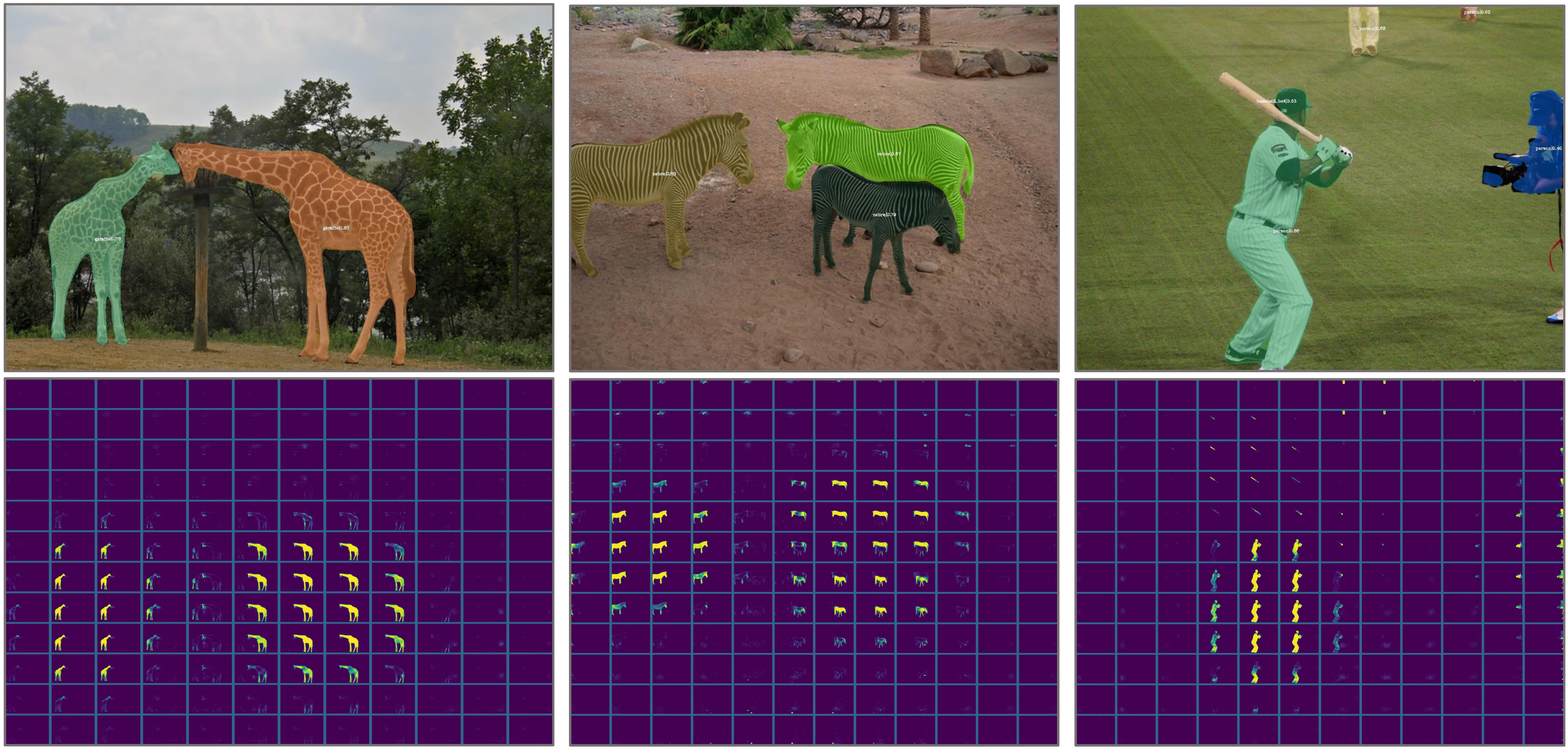}
\end{center}
   \caption{\textbf{\OurMethod behavior.} We show the visualization of soft mask prediction.
   Here $S=12$.
   For each column, the top one is the instance segmentation result, and the bottom one shows the mask activation maps.
   The sub-figure $(i, j)$ in an activation map indicates the mask prediction results  (after zooming out) generated by the corresponding mask channel. %
   }
\label{fig:vis_softmasks}
\end{figure*}

\subsection{Main Results}

We compare \OurMethod to the state-of-the-art methods in instance segmentation on MS COCO \texttt{test}-\texttt{dev} in Table~\ref{tab:sota}. \OurMethod with ResNet-101 achieves a mask AP of 37.8\%, the state of the art among existing \emph{two-stage} instance segmentation methods such as Mask R-CNN. \OurMethod outperforms all previous \emph{one-stage} methods, including TensorMask~\cite{Chen_2019_ICCV}.
Some \OurMethod outputs are visualized in Figure~\ref{fig:some_vis}, and more examples are in the supplementary.

\subsection{How \OurMethod Works?}
We show the network outputs generated by
$S = 12$
grids (Figure~\ref{fig:vis_softmasks}). The sub-figure $(i, j)$ indicates the soft mask prediction results generated by the corresponding mask channel. Here we can see that different instances activates at different mask prediction channels. By explicitly segmenting instances at different positions, \OurMethod converts the instance segmentation problem into a position-aware classification task. Only one instance will be activated at each grid, and one instance may be predicted by multiple adjacent mask channels. During inference, we use NMS to suppress these redundant masks.

\subsection{Ablation Experiments}
\label{subsec:ablation}

\myparagraph{Grid number.}
We compare the impacts of grid number on the performance with single output feature map as shown in Table~\ref{tab:numgrid}. The feature is generated by merging C3, C4, and C5 outputs in ResNet (stride: 8). To our surprise, $S = 12$
can already achieve
27.2\%  AP on the challenging MS COCO dataset. \OurMethod achieves 29\%  AP when improving the grid number to 24. This results indicate that our single-scale \OurMethod
can be applicable to some scenarios where object scales do not vary much.

\begin{table}[t]
    \centering
     \caption{The impact of \textbf{grid number and FPN}.
 FPN
 significantly improves the performance
 thanks to its ability to deal with varying sizes of objects.
 }
    \scalebox{0.99}{
    \begin{tabular}{c|ccc|ccc}
        grid number &AP & AP$_{50}$ & AP$_{75}$ & AP$_{S}$ & AP$_{M}$ & AP$_{L}$\\
        \Xhline{1pt}
        12&  27.2 & 44.9  & 27.6  & 8.7  & 27.6  & 44.5   \\
        24&  29.0  & 47.3  & 29.9  & 10.0  & 30.1  & 45.8    \\
        36&   28.6  & 46.3  & 29.7  & 9.5  & 29.5  & 45.2   \\
        Pyramid & 35.8  & 57.1  & 37.8  & 15.0  & 38.7  & 53.6  \\

    \end{tabular}}

     \label{tab:numgrid}
\end{table}

\myparagraph{Multi-level Prediction.}
From Table~\ref{tab:numgrid} we can see that our single-scale \method
could already get 29.0 AP on MS COCO dataset. In this ablation, we show that the performance could be further improved via multi-level prediction using FPN~\cite{fpn}.
We use five pyramids to segment objects of different scales (details in supplementary).  Scales of ground-truth masks are explicitly used to assign them to the levels of the pyramid.
From P2 to P6, the corresponding grid numbers are $[40, 36, 24, 16, 12]$ respectively.
Based on our multi-level prediction, we further achieve 35.8 AP. As expected, the performance over all the metrics has been largely improved.

%
\iffalse
\begin{table}[htbp]
    \centering
     \caption{we use five \textbf{FPN pyramids} to segment objects of different scales. The grid number increases for smaller instances due to larger existence space. }
    \scalebox{0.9}{
    \begin{tabular}{l|ccccc}
        pyramid & P2 & P3 & P4 & P5 & P6\\
        \Xhline{1pt}
        re-scaled stride & 8 & 8 & 16 & 32 & 32\\
        \hline
        grid number & 40 & 36 & 24 & 16 & 12\\
        \hline
        instance scale &$<$96 & 48$\sim$192 & 96$\sim$384 & 192$\sim$768 & $\geq$384\\
    \end{tabular}}

     \label{tab:fpn_seg}
\end{table}
\fi

\myparagraph{CoordConv.} Another important component that facilitates our \method paradigm is the \textit{spatially variant} convolution (CoordConv~\cite{coordconv}). As shown in Table~\ref{tab:coordconv}, the
standard convolution can already have spatial variant property to some extent, which is in accordance with the observation in~\cite{coordconv}.
As also revealed in~\cite{Islam2020How}, CNNs can implicitly learn the absolute position information from the commonly used zero-padding operation.
However, the implicitly learned position information is coarse and inaccurate.
When making the convolution access to its own input coordinates through
concatenating extra coordinate channels,
our method enjoys 3.6 absolute AP gains.
Two or more CoordConvs do not bring
noticeable
improvement.
It suggests that a single CoordConv already enables the predictions to be well spatially variant/position sensitive.

\begin{table}[h]
    \centering
     \caption{\textbf{Conv vs.\  CoordConv.}
    CoordConv can considerably improve AP upon standard convolution.
    Two or more layers of CoordConv are not necessary.
 }
    \scalebox{0.955}{
    \begin{tabular}{c|ccc|ccc}
        \#CoordConv &AP & AP$_{50}$ & AP$_{75}$ & AP$_{S}$ & AP$_{M}$ & AP$_{L}$\\
        \Xhline{1pt}
        0&  32.2 & 52.6 & 33.7  & 11.5  & 34.3   & 51.6    \\
        1&  35.8  & 57.1  & 37.8  & 15.0  & 38.7  & 53.6   \\
        2&  35.7 & 57.0 & 37.7 & 14.9 & 38.7 & 53.3   \\
        3&  35.8 & 57.4 & 37.7 & 15.7 & 39.0 & 53.0   \\
    \end{tabular}}

     \label{tab:coordconv}
\end{table}

\myparagraph{Loss function.} Table \ref{tab:loss} compares different loss functions for our mask optimization branch. The methods include conventional Binary Cross Entropy (BCE), Focal Loss (FL), and Dice Loss (DL).
To obtain improved performance, for Binary Cross Entropy we set a mask loss weight of 10 and a pixel weight of 2 for positive samples.
The mask loss weight of Focal Loss is set to 20.
As shown, the Focal Loss works much better than ordinary Binary Cross Entropy loss.
It is because that the majority of pixels of an instance mask are in background, and the Focal Loss is
designed to mitigate
the sample imbalance problem by decreasing the loss of well-classified samples.
However, the Dice Loss achieves the best results without the need of manually adjusting the loss hyper-parameters.
Dice Loss
views the pixels as a whole object and could establish
the right balance between foreground and background pixels automatically.
Note that with carefully tuning the balance hyper-parameters and introducing
other training tricks,  the results of Binary Cross Entropy
and Focal Loss may be considerably improved. However the point here is that with
the
Dice Loss, training  typically becomes much more stable and more likely to attain good results without using much heuristics.
To make a fair comparison, we also show the results of Mask R-CNN with Dice loss in the supplementary, which performs worse (-0.9AP) than original BCE loss.

\begin{table}[t]
    \centering
     \caption{\textbf{Different loss functions}
 may be employed
 in the mask branch. The Dice loss (DL) leads to best AP and is more stable to train.}
    \scalebox{0.95}{
    \begin{tabular}{c|ccc|ccc}
        mask loss &AP & AP$_{50}$ & AP$_{75}$ & AP$_{S}$ & AP$_{M}$ & AP$_{L}$\\
        \Xhline{1pt}
        BCE& 30.0 & 50.4  & 31.0  & 10.1  & 32.5  & 47.7   \\
        FL& 31.6 & 51.1 & 33.3 & 9.9 & 34.9 & 49.8  \\
        DL&  35.8  & 57.1  & 37.8  & 15.0  & 38.7  & 53.6   \\
    \end{tabular}}

     \label{tab:loss}
\end{table}

\myparagraph{Alignment in the category branch.}
In the category prediction branch, we must match the convolutional features with spatial size $H$$\times$$W$ to $S$$\times$$S$. Here, we compare three common implementations: interpolation, adaptive-pool, and region-grid-interpolation. (a) Interpolation: directly bilinear interpolating to the target grid size; (b) Adaptive-pool: applying a 2D adaptive max-pool over $H$$\times$$W$ to $S$$\times$$S$; (c) Region-grid-interpolation: for each grid cell, we use bilinear interpolation conditioned on dense sample points, and aggregate the results with average.
\iffalse
\begin{itemize}
\itemsep -0.0051cm
    \item Interpolation: Directly bilinear interpolating to the target grid size;
    \item Adaptive-pool: Applying a 2D adaptive max-pool over $H$$\times$$W$ to $S$$\times$$S$;
    \item Region-grid-interpolation: For each grid cell, we use bilinear interpolation conditioned on dense sample points, and aggregate the results with average.
\end{itemize}
\fi
From our observation, there is no noticeable performance gap between these variants ($\pm$ 0.1AP), indicating that the alignment process
does not have a significant impact on the final accuracy.

\myparagraph{Different head depth.}
In \OurMethod, instance segmentation is formulated as a pixel-to-pixel task and we exploit the spatial layout of masks by using an FCN. In Table~\ref{tab:depth_ap}, we compare different head depth used in our work. Changing the head depth from 4 to 7 gives 1.2 AP gains.  The results show that when the depth grows beyond 7, the performance becomes stable. In this paper, we  use depth being 7 in other experiments.

\begin{table}[h]
    \centering
     \caption{\textbf{Different head depth.} We  use depth being 7 in other experiments, as the performance becomes stable when the depth grows beyond 7.}
    \scalebox{0.94}{
    \begin{tabular}{c|ccccc}
        head depth & 4 & 5 & 6 & 7 & 8 \\
        \Xhline{1pt}
        AP & ~34.6~ & ~35.2~ & ~35.5~ & ~35.8~ & ~35.8~  \\
    \end{tabular}}
     \label{tab:depth_ap}
\end{table}

\iffalse
\begin{figure}[t]
\centering
    \includegraphics[width=0.5\linewidth]{figures/depth_ap.pdf}
   \caption{Results on the COCO \texttt{val2017} set using \textbf{different head depth} on ResNet-50-FPN.}
\label{fig:depth_ap}
\end{figure}
\fi

Previous works (\eg,  Mask R-CNN) usually adopt four conv layers for mask prediction. In \OurMethod, the mask is conditioned on the spatial position and we simply attach the coordinate to the beginning of the head. The mask head must have enough representation power to learn such transformation. For the semantic category branch, the computational overhead is negligible since $S^2 \ll H\times W$.

\iffalse
\textbf{Different backbones.}
Table~\ref{tab:backbone} shows \method with various backbones. These models are all trained using 24 epochs, and tested on the val2017 dataset. Our method benefits from deeper networks (50 vs.\  101) and advanced design (ResNeXt~\cite{resnext}).
\begin{table}[h]
    \centering
    \scalebox{0.78}{
    \begin{tabular}{r |ccc|ccc|c}
        backbone &AP & AP$_{50}$ & AP$_{75}$ & AP$_{S}$ & AP$_{M}$ & AP$_{L}$ & FPS\\
        \Xhline{1pt}
        ResNet-50     &  32.4 & 54.1 & 33.4 & 13.5 & 34.7 & 49.0 & 10.4 \\
        ResNet-101    &  33.7 & 55.7 & 34.7 & 14.1 & 36.3 & 51.8 & 9.2 \\
        ResNeXt-101   & 34.9 & 57.5 & 36.4 & 15.1 & 37.4 & 53.6 & 7.9 \\
    \end{tabular}}
 \caption{\textbf{Backbone architecture}.
 %
 Backbones with a larger capacity
 show expected gains, and ResNeXt improves on ResNet.}
     \label{tab:backbone}
\end{table}
\fi

\subsection{\OurMethod-512}
Speed-wise, the Res-101-FPN SOLO runs at
10.4 FPS on a V100 GPU (all post-processing included), vs.
TensorMask’s 2.6 FPS and Mask R-CNN's 11.1 FPS.
We also train a smaller version of \OurMethod designed to speed up the inference. We use a model with smaller input resolution (shorter image size of 512 instead of 800). Other training and testing parameters are the
same between \OurMethod-512 and \OurMethod.

\begin{table}[h]
    \centering
     \caption{\textbf{\OurMethod-512}. \OurMethod-512 uses a model with smaller input size. All models are evaluated on \texttt{val2017}. Here the models are trained with ``6$\times$'' schedule.}
    \scalebox{0.9985}{
    \begin{tabular}{c | l |ccc|ccc|c}
         & backbone & AP & AP$_{50}$ & AP$_{75}$ & fps \\
        \Xhline{1pt}
        \OurMethod & ResNet-50-FPN &  36.0 & 57.5 & 38.0 & 12.1   \\
        \OurMethod & ResNet-101-FPN &  37.1 & 58.7 & 39.4 & 10.4  \\
        \hline
        \OurMethod-512 & ResNet-50-FPN &  34.2 & 55.9 & 36.0 & 22.5 \\
        \OurMethod-512 & ResNet-101-FPN &  35.0 & 57.1 & 37.0 & 19.2 \\
    \end{tabular}}

     \label{tab:fast_version}
\end{table}
With 34.2 mask AP, \OurMethod-512
achieves
a model inference speed of 22.5 FPS, showing that \OurMethod has potentiality for real-time instance segmentation applications. The speed is reported on a single V100 GPU by averaging 5 runs.

\subsection{Error Analysis}
To quantitatively understand \OurMethod for mask prediction, we perform an error analysis by replacing the predicted masks with ground-truth values.
For each predicted binary mask, we compute IoUs with ground-truth masks, and replace it with the most overlapping ground-truth mask. As reported in Table~\ref{tab:error_analysis}, if we replace the predicted masks with ground-truth masks, the AP increases to 68.1\%. This experiment suggests that there are still ample room for improving the mask branch.
We expect techniques developed (a) in semantic segmentation, and (b) for dealing occluded/tiny objects could be applied to boost the performance.

\begin{table}[h]
    \centering
     \caption{\textbf{Error analysis}. Replacing the predicted instance mask with the ground-truth ones improves the mask AP from 37.1 to 68.1, suggesting that the mask branch still has ample room to be improved. The models are based on ResNet-101-FPN.}
    \scalebox{0.995}{
    \begin{tabular}{r |ccc|ccc}
         &AP & AP$_{50}$ & AP$_{75}$ & AP$_{S}$ & AP$_{M}$ & AP$_{L}$\\
        \Xhline{1pt}
        baseline       & 37.1 & 58.7 & 39.4 & 16.0 & 41.1 & 54.2 \\
        w/gt mask     & 68.1 & 68.3 & 68.2 & 46.1 & 75.0 & 78.5 \\
    \end{tabular}}

     \label{tab:error_analysis}
\end{table}

\section{Decoupled \OurMethod}
\label{sec:decouple}
\begin{figure}[t]
\centering
\subfigure[Vanilla head]{
\includegraphics[width=0.36\textwidth]{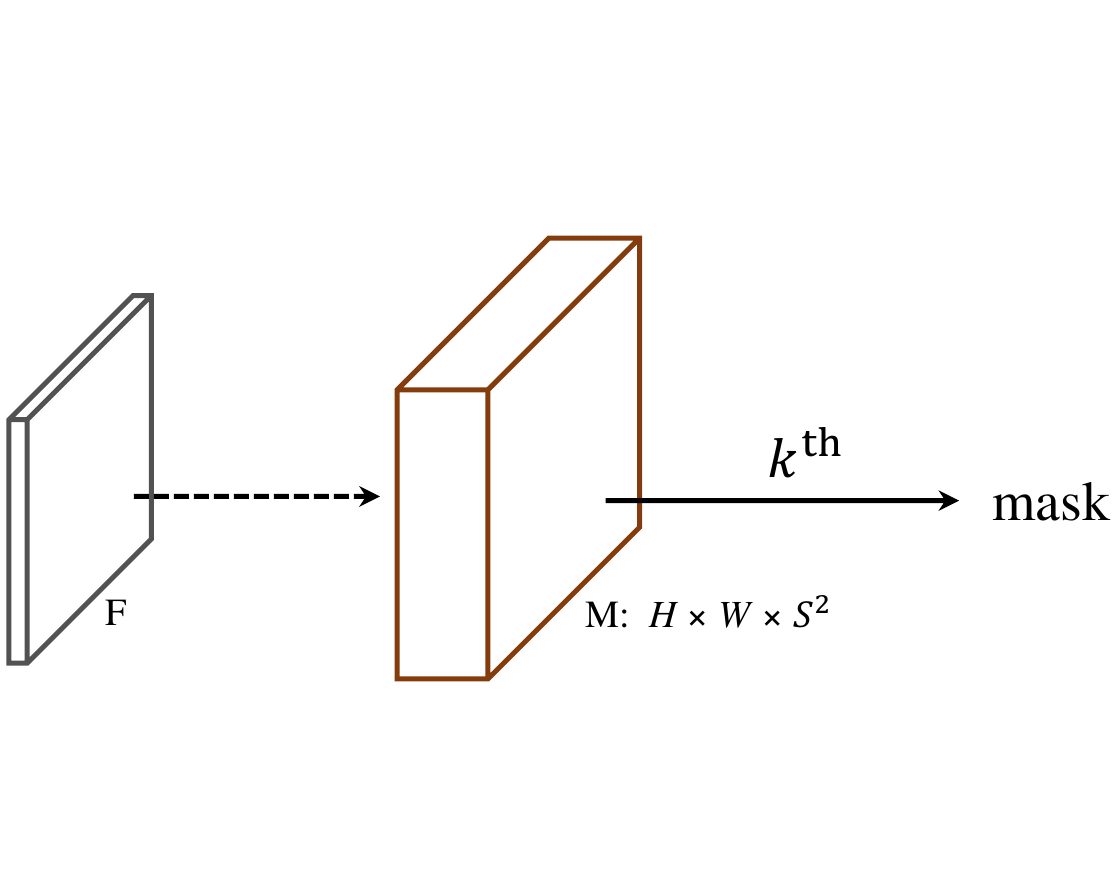}
\label{fig:decoupled_head_1}
}
\subfigure[Decoupled head]{
\includegraphics[width=0.36\textwidth]{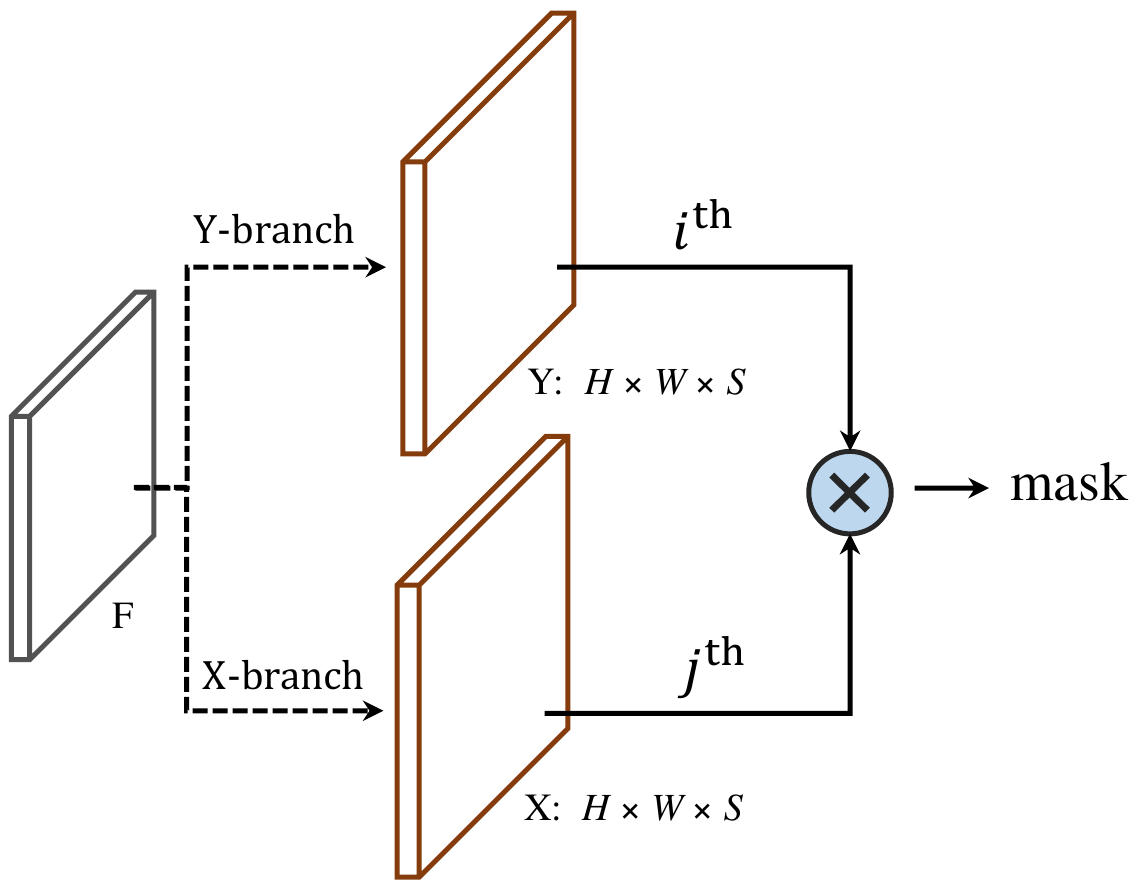}
\label{fig:decoupled_head_3}
}
\caption{\textbf{Decoupled \OurMethod head}. $F$ is input feature. Dashed arrows denote convolutions. $k=i \cdot  S + j$. `$\otimes$' denotes element-wise multiplication.  }
\label{fig:decoupled_head}
\end{figure}

Given an predefined grid number, \eg, $S=20$, our \OurMethod head outputs $S^2 = 400$ channel maps.
However, the prediction is somewhat redundant as in most cases the objects are located sparsely in the image.
In this section, we further introduce an equivalent and significantly more efficient variant of the vanilla \OurMethod, termed \textbf{Decoupled \OurMethod}, shown in Figure~\ref{fig:decoupled_head}.

In Decoupled \OurMethod, the original output tensor $M\in \mathbb{ R}^{H\times W\times S^2}$  is replaced with two output tensors $X\in \mathbb{ R}^{H\times W\times S}$  and $Y\in \mathbb{ R}^{H\times W\times S}$, corresponding two axes respectively.
Thus, the output space is decreased from  $H$$\times$$W$$\times$$S^2$ to  $H$$\times$$W$$\times$$2S$.
For an object located at grid location $(i, j)$,
the mask prediction of that object is defined as the element-wise multiplication of two channel maps:
\begin{equation}
\begin{aligned}
\mathbf{m}_k = \mathbf{x}_j \otimes \mathbf{y}_i,
\end{aligned}
\end{equation}
where $\mathbf{x}_j$ and $\mathbf{y}_i$ are the $j^{th}$ and $i^{th}$ channel map of $X$ and $Y$ after \texttt{sigmoid} operation.
The motivation behind this is that the probability of a pixel belonging to location category $(i, j)$ is the joint probability of belonging to $i^{th}$ row and $j^{th}$ column,
as the horizontal and vertical location categories are independent.
%

\iffalse
\begin{table}
    \centering
     \caption{Vanilla head vs.\  Decoupled head. The models are trained with``3$\times$'' schedule and evaluated on \texttt{val2017}. }
    \scalebox{0.85}{
    \begin{tabular}{r |ccc|ccc}
         &AP & AP$_{50}$ & AP$_{75}$ & AP$_{S}$ & AP$_{M}$ & AP$_{L}$\\
        \Xhline{1pt}
        Vanilla \OurMethod & 35.8  & 57.1  & 37.8  & 15.0  & 38.7  & 53.6   \\
        Decoupled \OurMethod & 35.8 & 57.2  & 37.7  & 16.3 & 39.1  & 52.2   \\
        %
    \end{tabular}}
     \label{tab:decoupled_solo}
\end{table}
\fi

%
We conduct experiments using the the same hyper-parameters as vanilla \OurMethod.
As shown in Table~\ref{tab:sota},
Decoupled \OurMethod even achieves slightly better performance (0.6 AP gains) than vanilla \OurMethod.
With DCN-101~\cite{dcn} backbone, we further achieve 40.5 AP, which is considerably better than current dominant approaches.
It indicates that the Decoupled \OurMethod serves as an efficient and equivalent variant in accuracy of \OurMethod.
Note that, as the output space is largely reduced, the Decoupled \OurMethod needs considerably less GPU memory during training and testing.

\def\visimgheight{2.97cm}
\def\visimgwidth{3.86cm}

\begin{figure*}[ht]
\centering 
{

\includegraphics[height=\visimgheight]{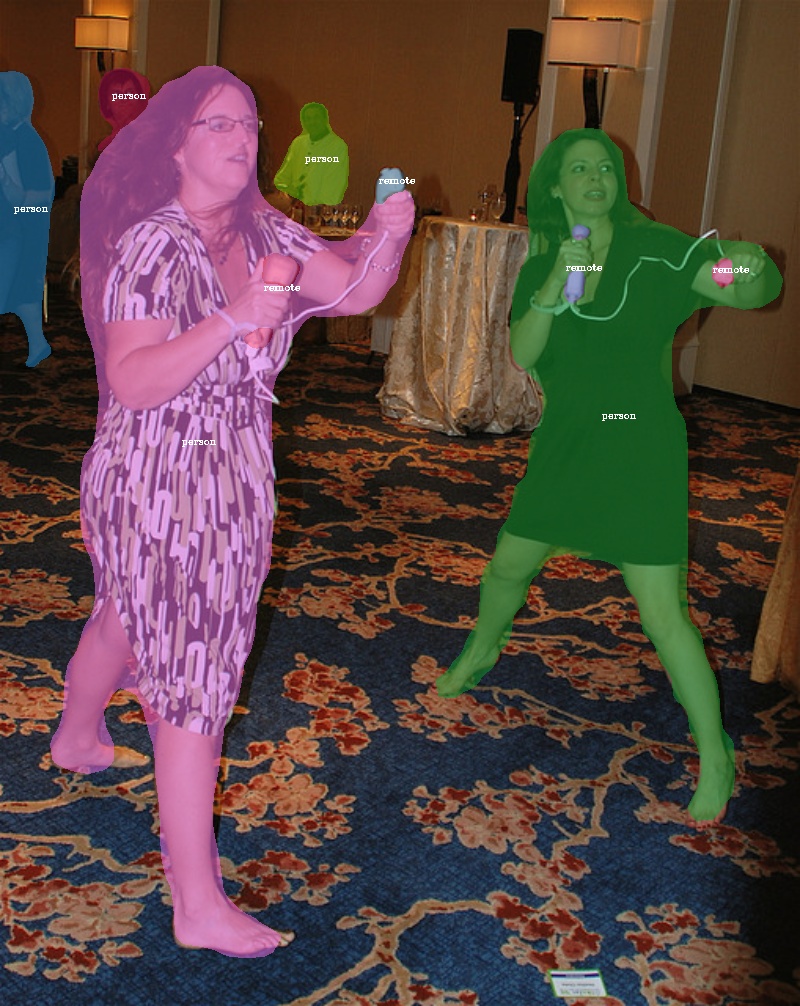}
\includegraphics[height=\visimgheight]{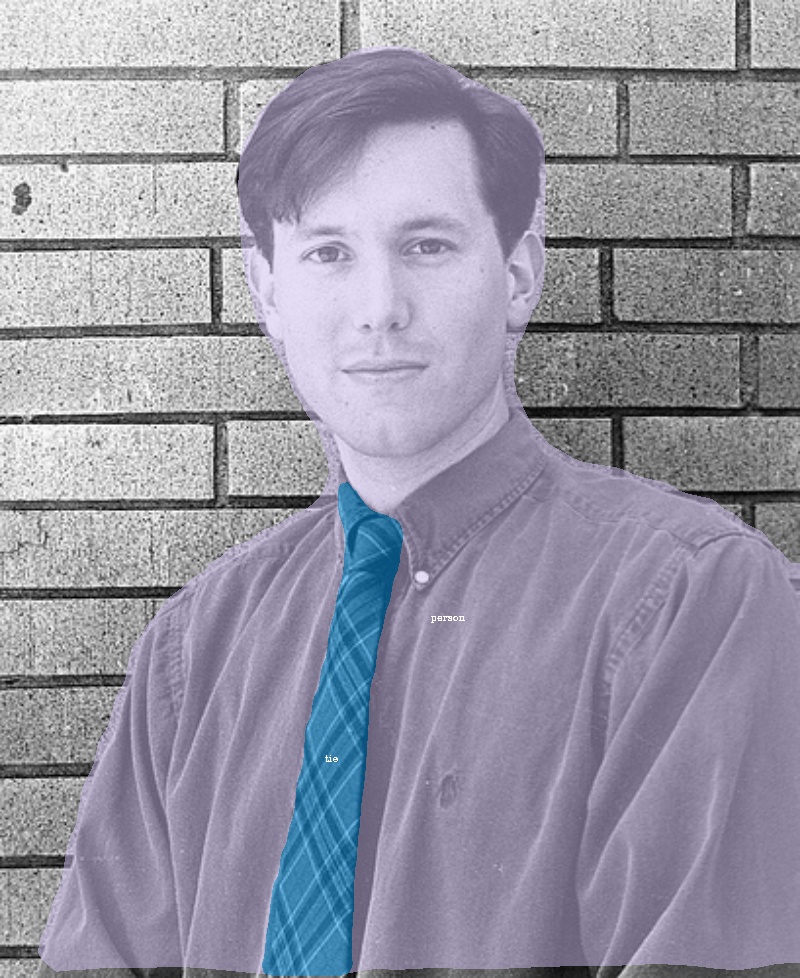}
\includegraphics[height=\visimgheight]{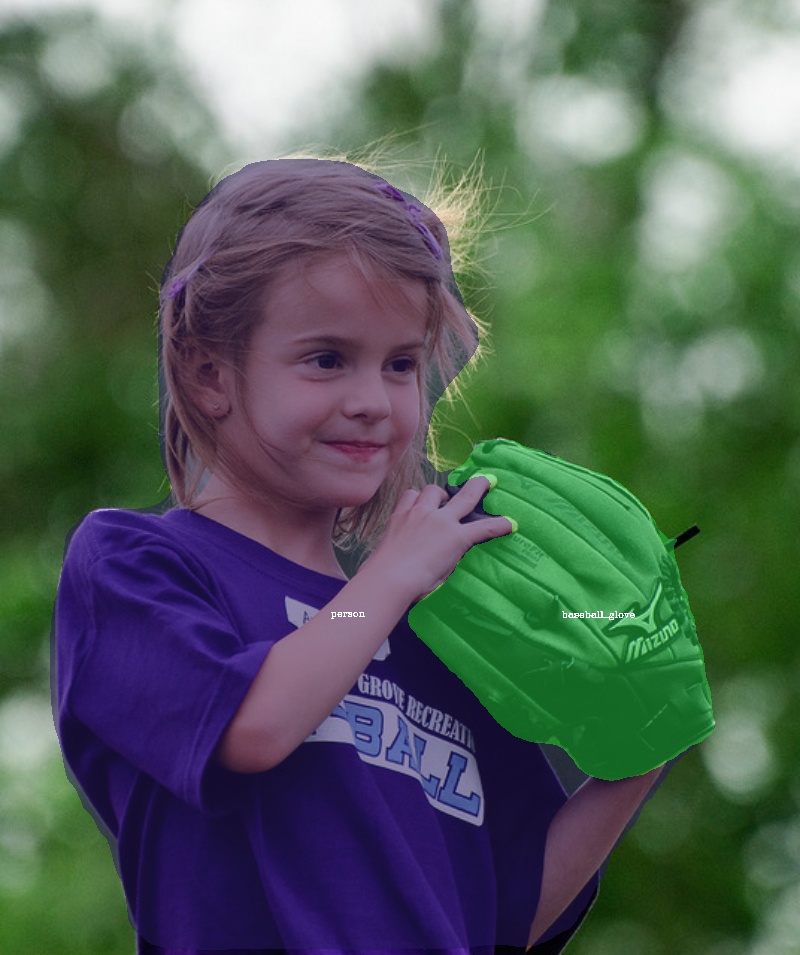}
\includegraphics[height=\visimgheight]{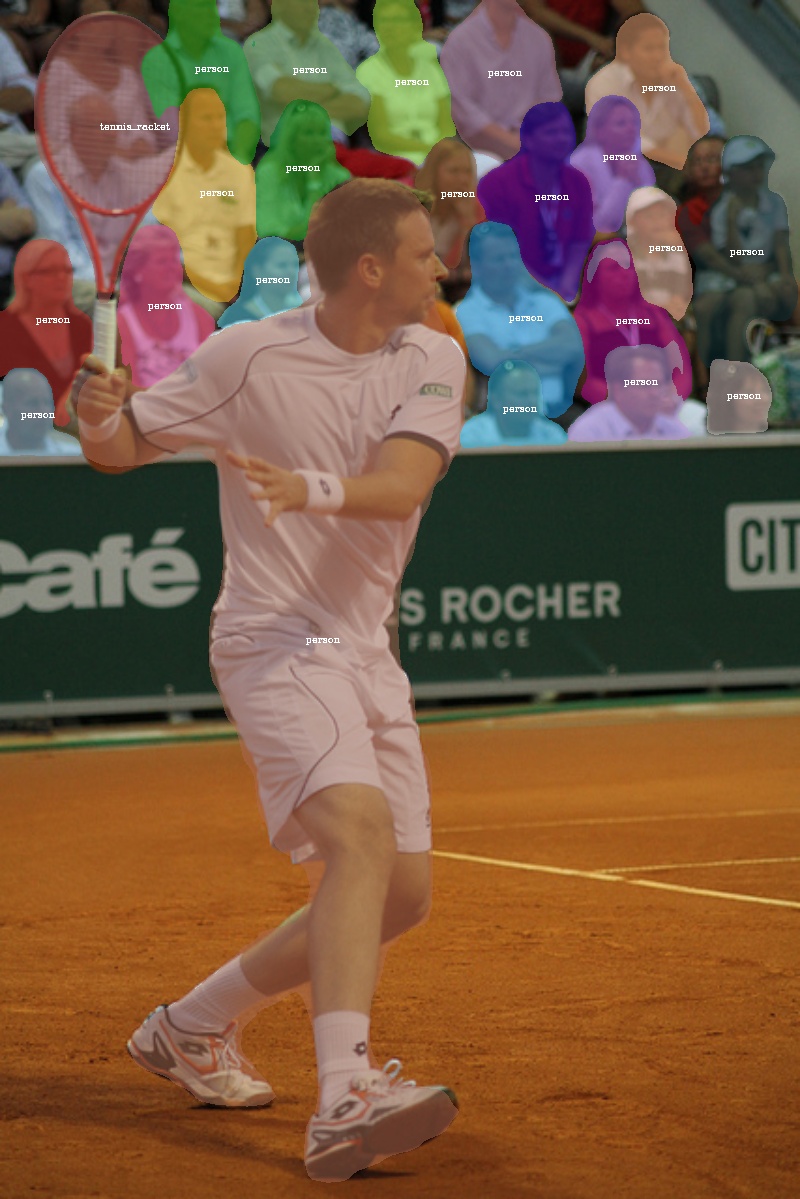}
\includegraphics[height=\visimgheight]{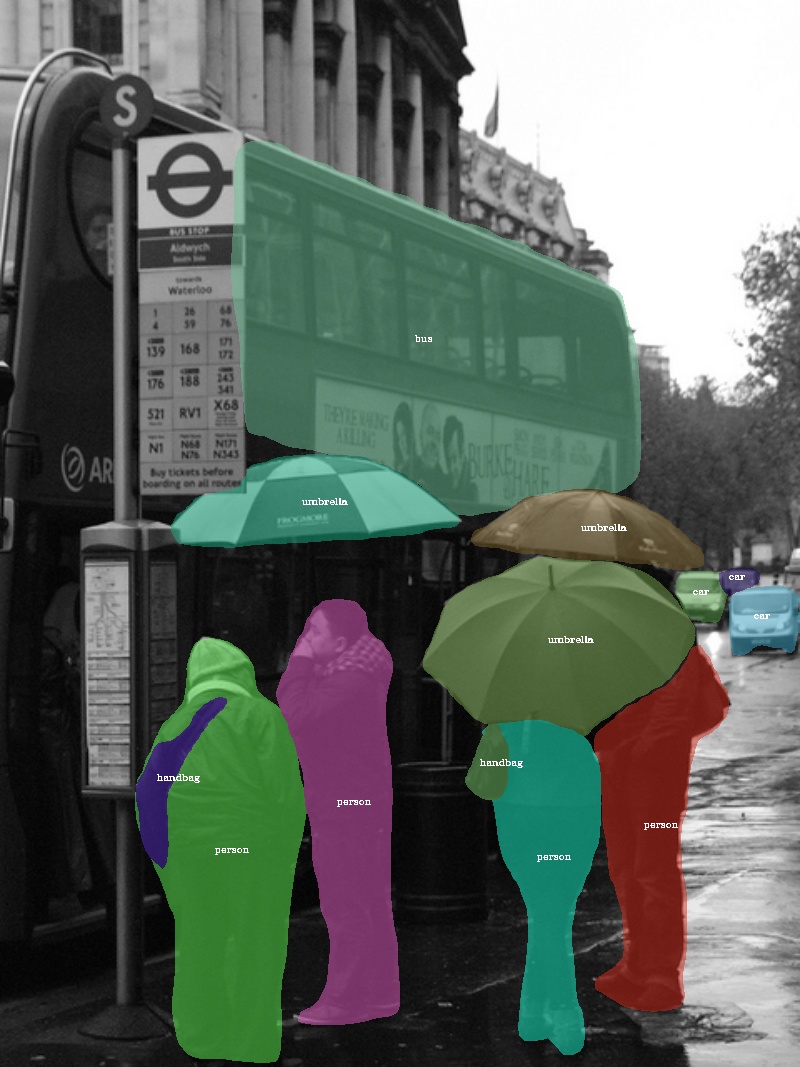}

\def\visimgheight2{2.65cm} 
\includegraphics[height=\visimgheight2]{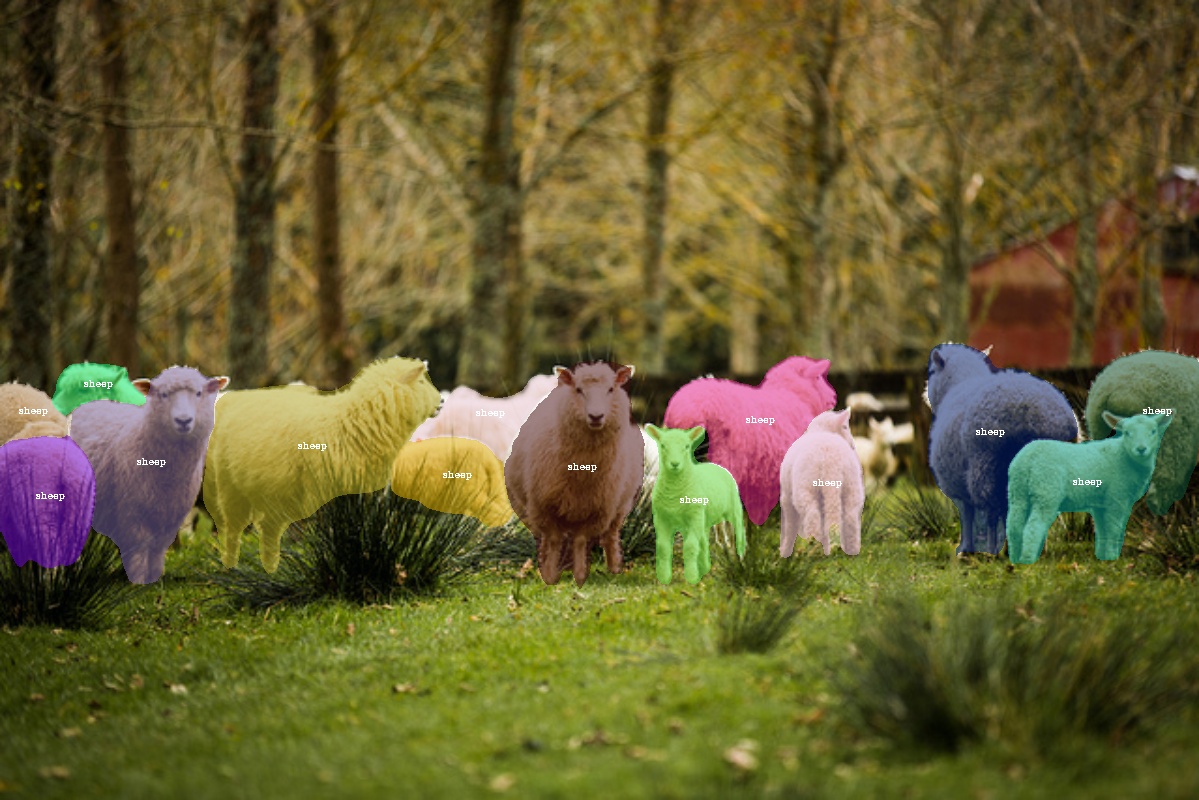}
\includegraphics[height=\visimgheight2]{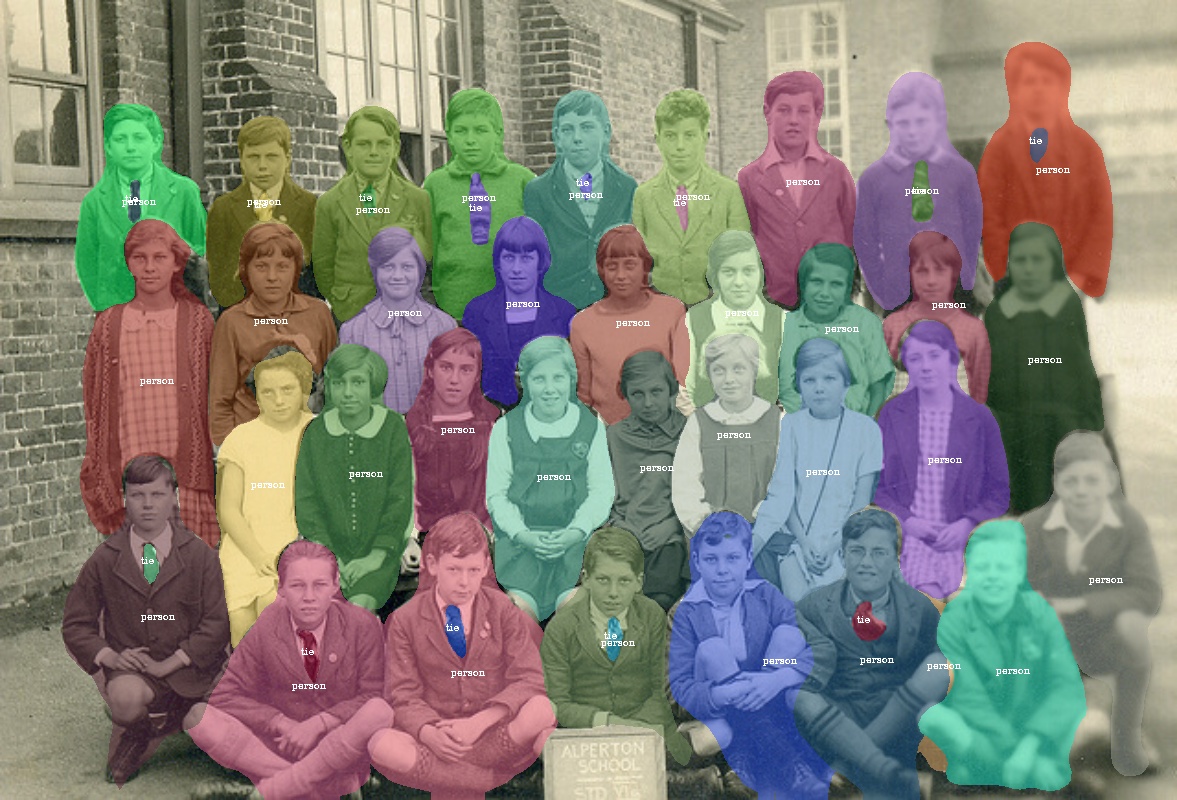}
\includegraphics[height=\visimgheight2]{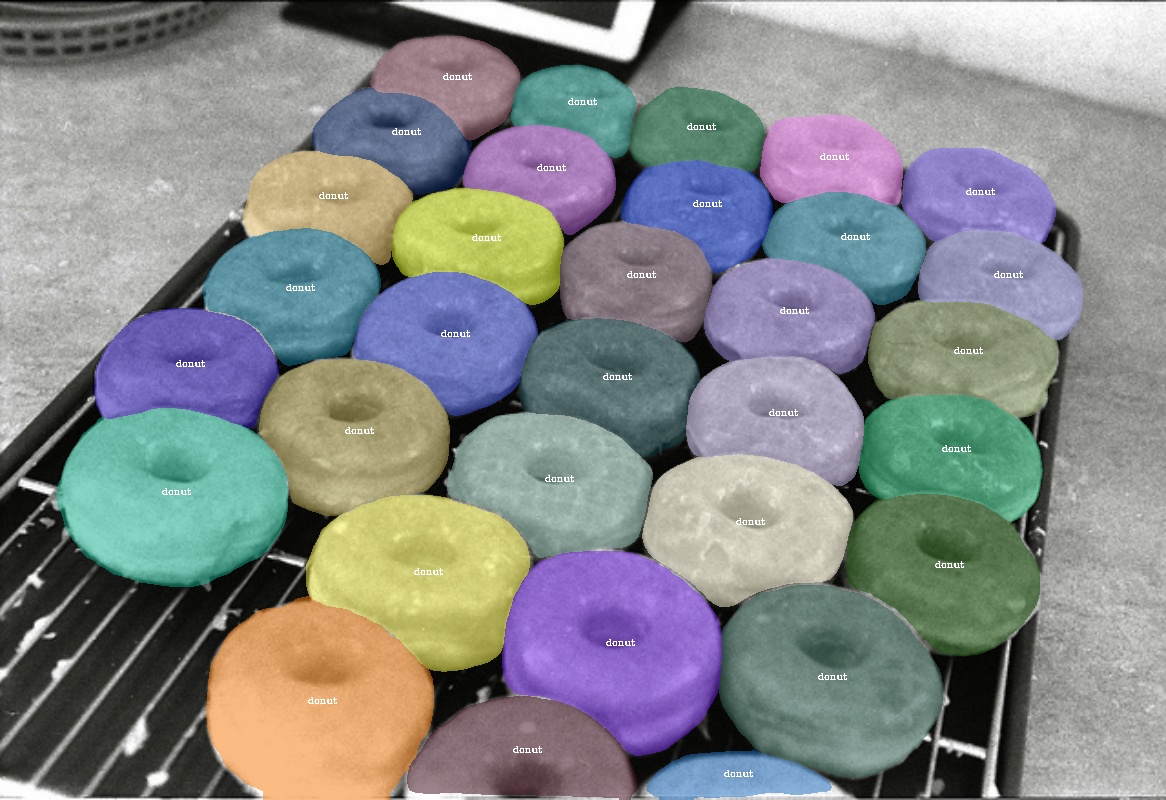}

} %
\caption{\textbf{Visualization of instance segmentation results}
using the Res-101-FPN backbone.
The model is trained on the COCO \texttt{train2017} dataset, achieving a mask AP of 37.8 on the COCO \texttt{test}-\texttt{dev}.
}
\label{fig:some_vis}
\end{figure*}

\section{Conclusion}
In this work we have developed a direct instance segmentation framework, termed \OurMethod.
Our  \OurMethod is end-to-end trainable and can directly map a raw input image to the desired instance masks with constant inference time, eliminating the need for the grouping post-processing as in bottom-up methods or the bounding-box detection and RoI operations in top-down approaches.
Given the simplicity, flexibility, and strong performance of \OurMethod,
we hope that
our \OurMethod can serve as a cornerstone for many instance-level recognition tasks.

\myparagraph{Acknowledgement} We would like to thank Dongdong Yu and Enze Xie for
the
discussion about
maskness and dice loss. We also thank Chong Xu and %
the ByteDance AI Lab team for technical support.
\bibliographystyle{splncs04}
\bibliography{CSRef}

\clearpage

\appendix
\section{More Method Details}

\iffalse
\subsection{Alignment in Category Branch}
In the category prediction branch, we need to match the convolutional features with spatial size $H$$\times$$W$ to $S$$\times$$S$. Here, we compare three common implementations: interpolation, adaptive-pool, and region-grid-interpolation.

\begin{itemize}
    \item Interpolation: Directly bilinear interpolating to the target grid size;
    \item Adaptive-pool: Applying a 2D adaptive max-pool over $H$$\times$$W$ to $S$$\times$$S$;
    \item Region-grid-interpolation: For each grid cell, we use bilinear interpolation conditioned on dense sample points, and aggregate the results with average.
\end{itemize}

From our observation, there is no noticeable performance gap between these variants ($\pm$ 0.1AP), indicating the alignment process is fairly flexible. In all our experiments, we use the bilinear interpolation choice.
\fi

\subsection{Multi-level Prediction}
We use five FPN pyramids to segment objects of different scales (Table~\ref{tab:fpn_seg}). Scales of ground-truth masks are explicitly used to assign them to the levels of the pyramid. Multi-level prediction gives 6.8 AP gains on the single-scale \OurMethod.

\begin{table}[htbp]
    \centering
     \caption{we use five \textbf{FPN pyramids} to segment objects of different scales. The grid number increases for smaller instances due to larger existence space. }
    \scalebox{0.9}{
    \begin{tabular}{l|ccccc}
        pyramid & P2 & P3 & P4 & P5 & P6\\
        \Xhline{1pt}
        re-scaled stride & 8 & 8 & 16 & 32 & 32\\
        \hline
        grid number & 40 & 36 & 24 & 16 & 12\\
        \hline
        instance scale &$<$96 & 48$\sim$192 & 96$\sim$384 & 192$\sim$768 & $\geq$384\\
    \end{tabular}}

     \label{tab:fpn_seg}
\end{table}

\section{More Experimental Results}

\subsection{Single-scale 1$\times$ Training}
We list the 1x single-scale results in Table~\ref{tab:singlescale_results}.

\begin{table}
    \centering
     \caption{Results with single-scale training. The models are trained with ``1$\times$'' schedule and evaluated on \texttt{val2017}. }
    \scalebox{0.85}{
    \begin{tabular}{r |ccc|ccc}
         &AP & AP$_{50}$ & AP$_{75}$ & AP$_{S}$ & AP$_{M}$ & AP$_{L}$\\
        \Xhline{1pt}
        \OurMethod  & 32.9 & 53.2 & 34.8 & 12.7 & 36.2 & 50.5  \\
        D-\OurMethod & 33.9 & 54.0 & 35.7 & 13.8 & 36.9 & 51.0 \\
    \end{tabular}}
     \label{tab:singlescale_results}
\end{table}

\subsection{Dice Loss}
To make a fair comparison, we show the results of Mask R-CNN with Dice loss in Table~\ref{tab:maskrcnn_dice}. It shows that Dice loss is not suitable for Mask R-CNN, as it performs worse (-0.9AP) than original BCE loss. It is because the 'detect-then-segment' methods do not have the fg/bg imbalance issue, as they segment the foreground pixels in a local bounding box.

\begin{table}
    \centering
     \caption{Mask R-CNN with Dice loss. The models are trained with``3$\times$'' schedule and evaluated on \texttt{val2017}. }
    \scalebox{0.85}{
    \begin{tabular}{r |ccc|ccc}
         &AP & AP$_{50}$ & AP$_{75}$ & AP$_{S}$ & AP$_{M}$ & AP$_{L}$\\
        \Xhline{1pt}
        BCE \ & 36.2 &	58.0 & 38.9 & 20.1 & 39.5 & 49.0  \\
        DL \ & 35.3 & 57.8 & 37.4 & 19.3 & 39.1 & 47.8  \\
    \end{tabular}}
     \label{tab:maskrcnn_dice}
\end{table}

\subsection{\method for Instance Contour Detection}
\begin{figure*}[ht]
\begin{center}
    \includegraphics[width=0.95\linewidth]{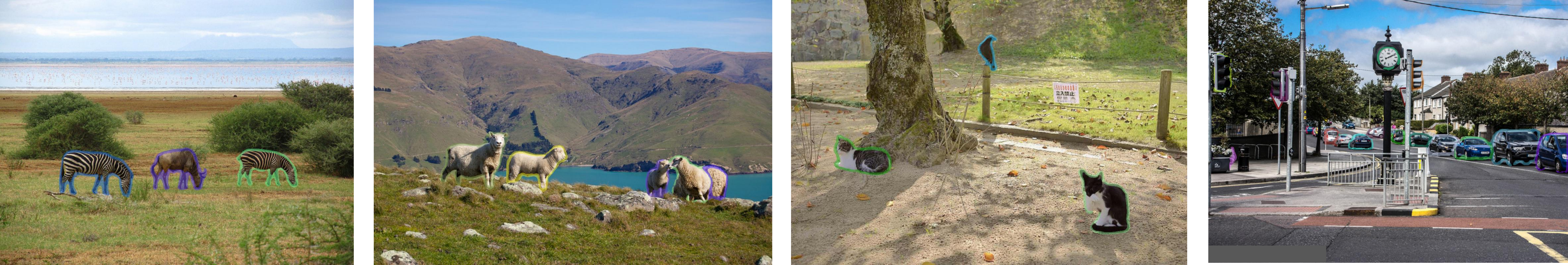}
\end{center}
   \caption{Visualization of \OurMethod for \textbf{instance contour detection}. The model is trained on COCO \texttt{train2017} dataset with ResNet-50-FPN. Each instance contour is shown in a
   different
   color.}
\label{fig:edge_det}
\end{figure*}

Our framework can easily be extended to instance contour detection.
We first convert the ground-truth masks in MS COCO into instance contours using OpenCV's \texttt{findContours} function, and then use the binary contours to optimize the mask branch in parallel with the semantic category branch. Here we use Focal Loss to optimize the contour detection, other settings are the same with the instance segmentation baseline. Figure~\ref{fig:edge_det} shows some contour detection examples generated by our model.  We provide these results as a proof of concept that
\method can be used in contour detection.

\subsection{Qualitative Results}
We show more visualization results in Figure~\ref{fig:Vis}.

\def\visimgheight{2.6cm}
\def\visimgwidth{3.2cm}

\begin{figure*}[ht]
\centering 
{

\def\visimgheighta{2.7cm}
\includegraphics[height=\visimgheighta]{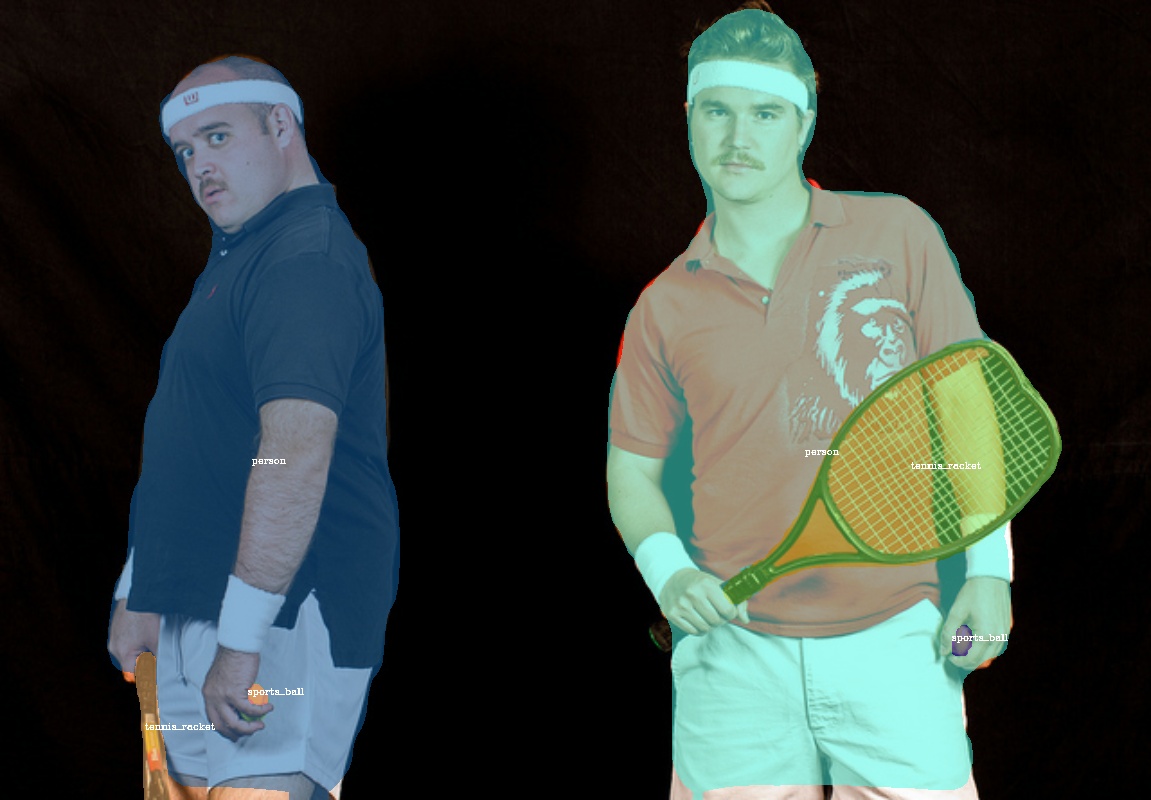}
\includegraphics[height=\visimgheighta]{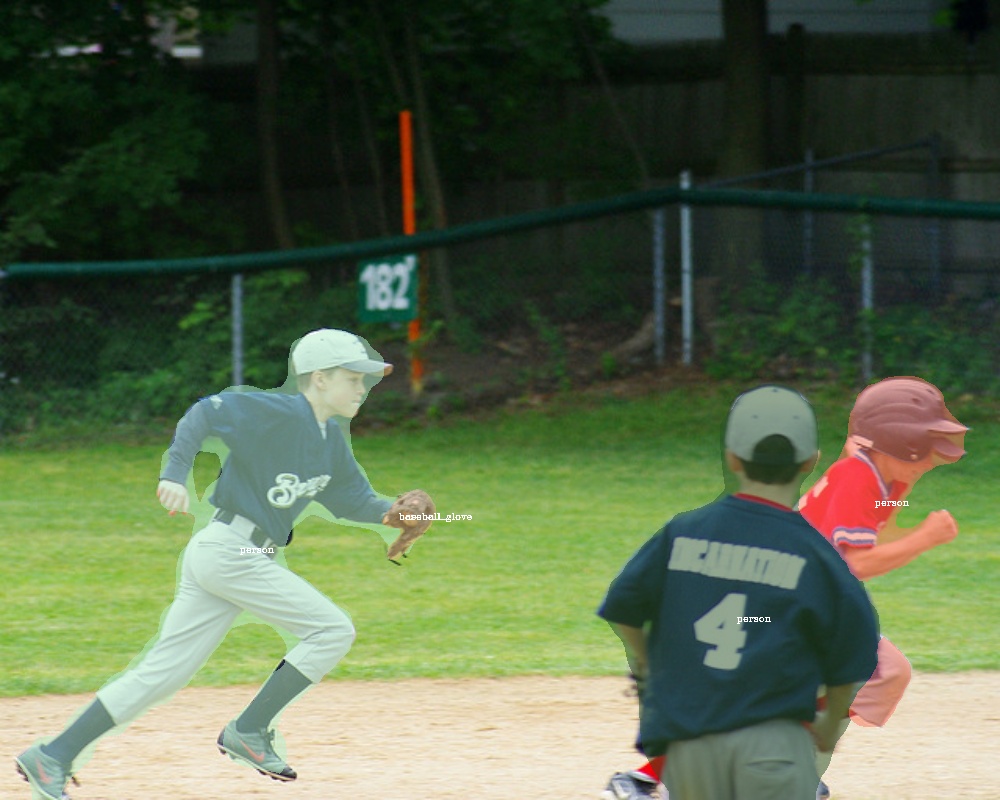}
\includegraphics[height=\visimgheighta]{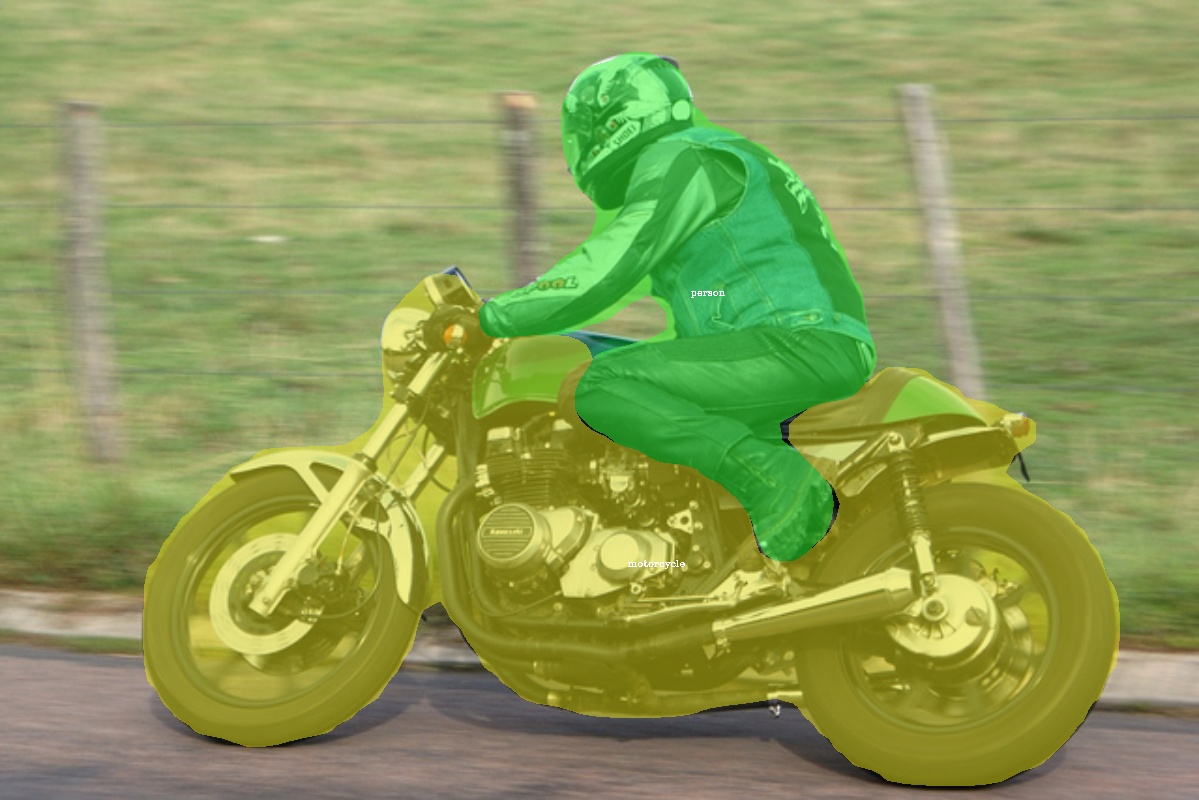}

\def\visimgheightb{2.67cm}
\includegraphics[width=\visimgwidth,height=\visimgheightb]{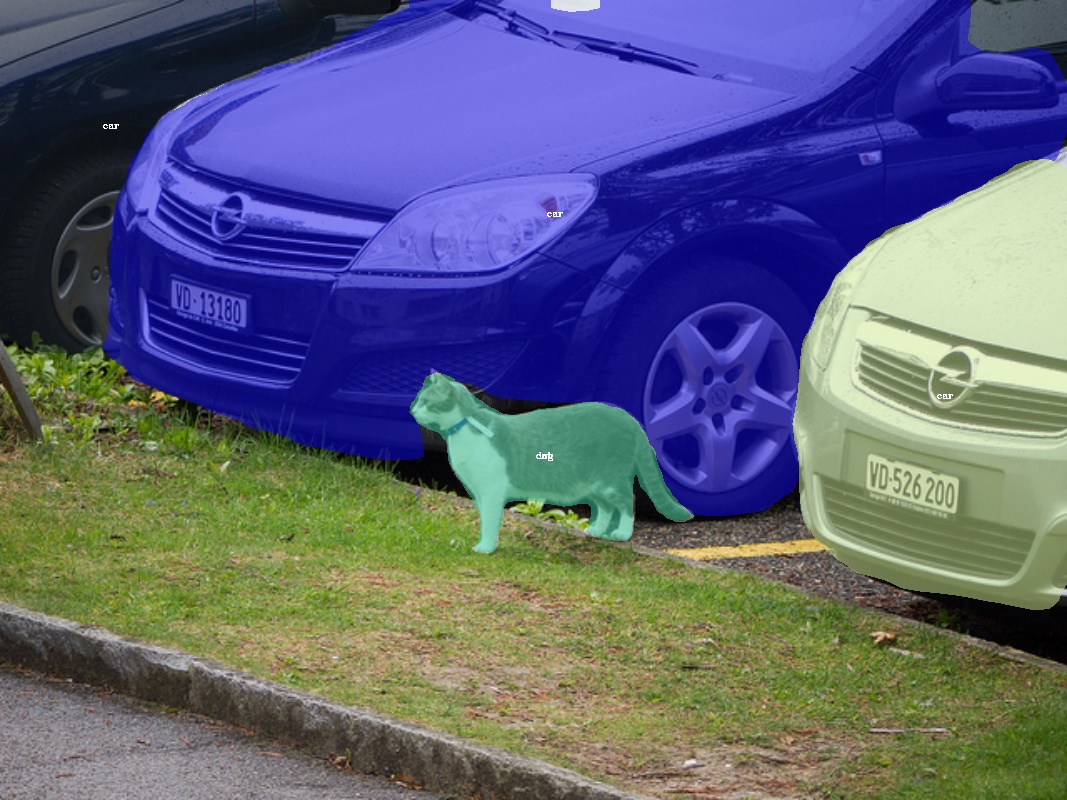}
\includegraphics[height=\visimgheightb]{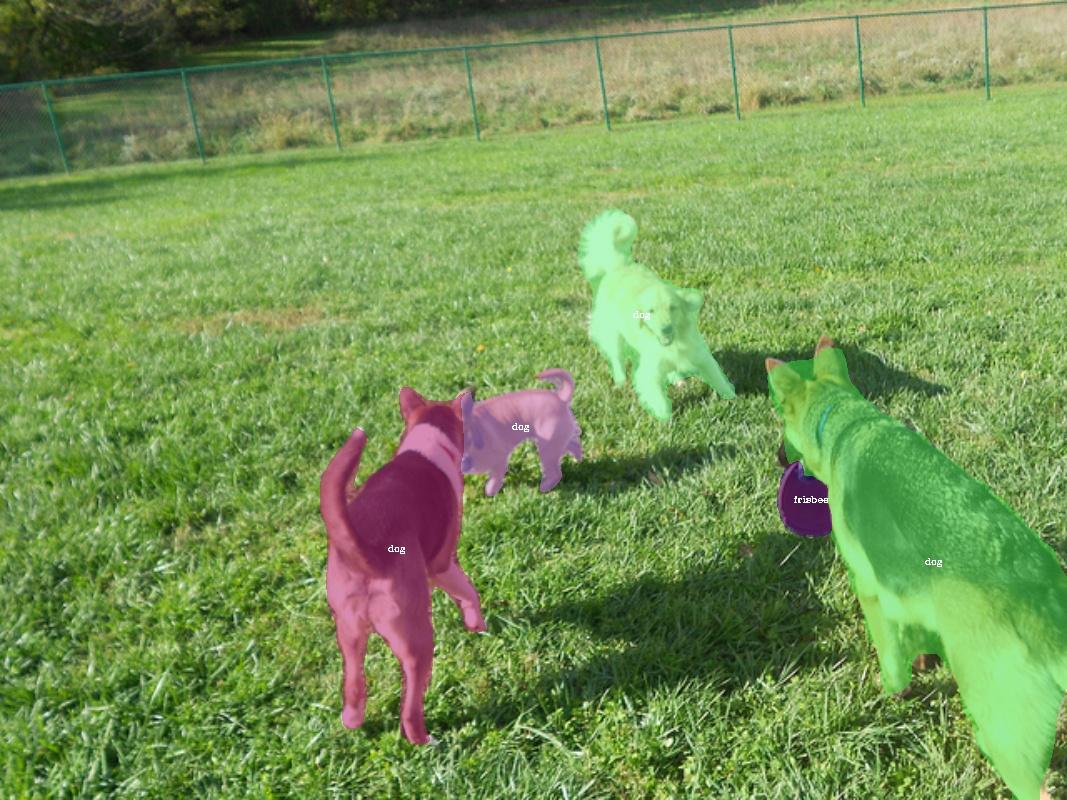}
\includegraphics[height=\visimgheightb]{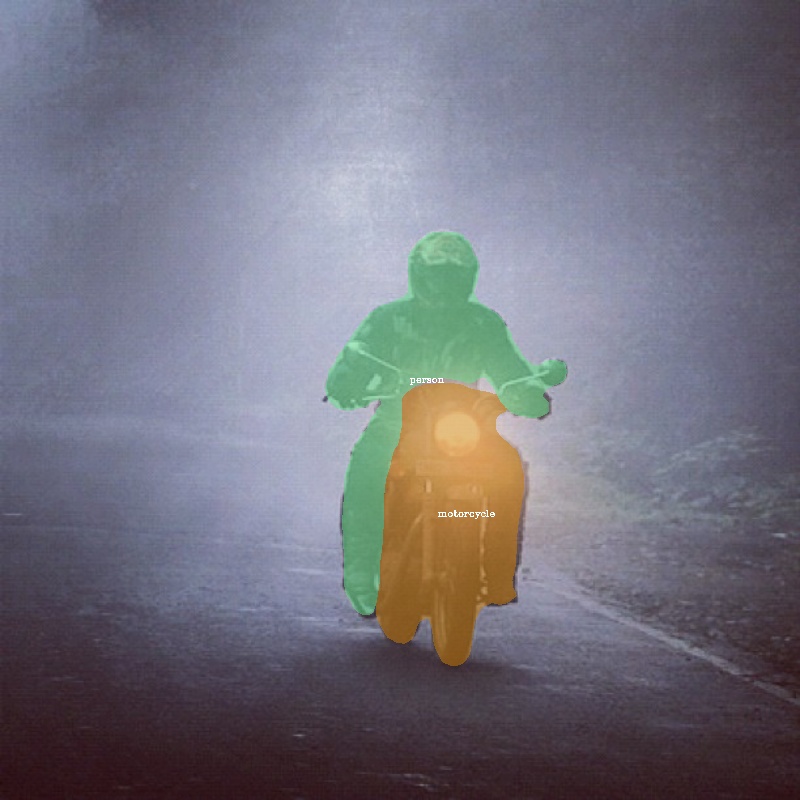}
\includegraphics[height=\visimgheightb]{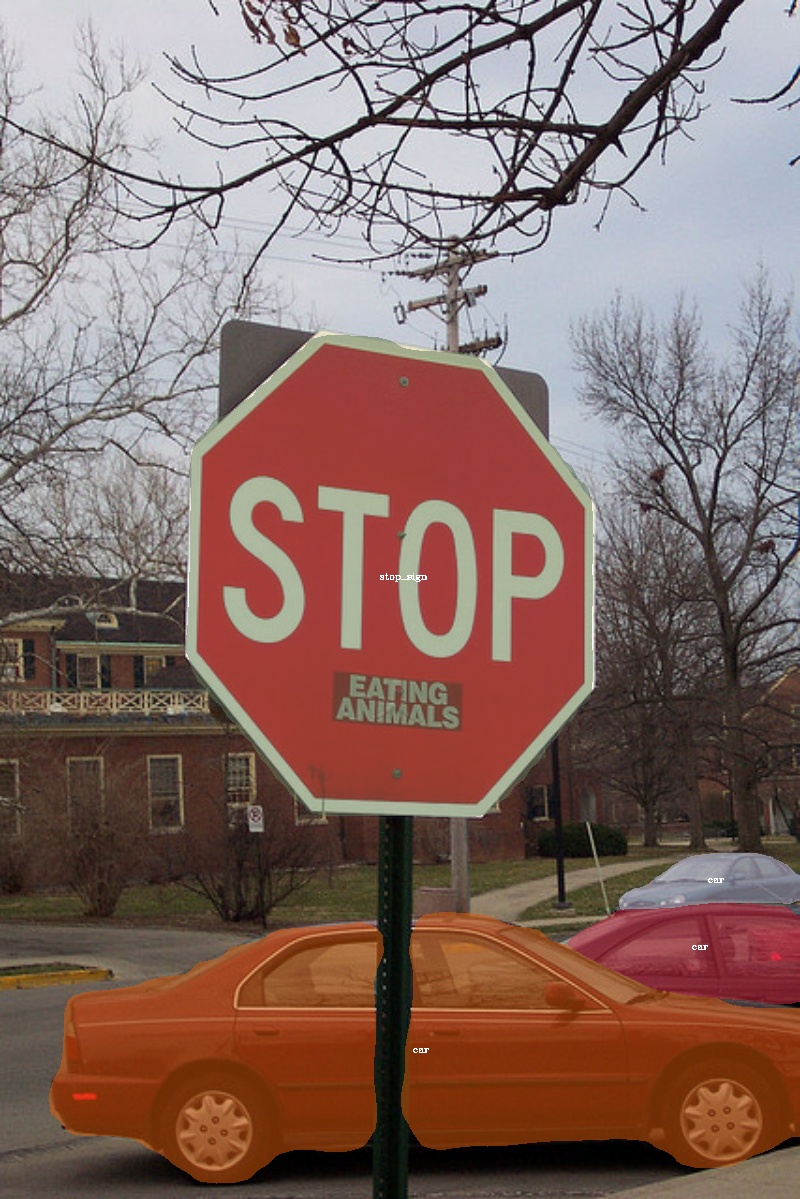}

\def\visimgheightc{2.57cm}
\includegraphics[height=\visimgheightc]{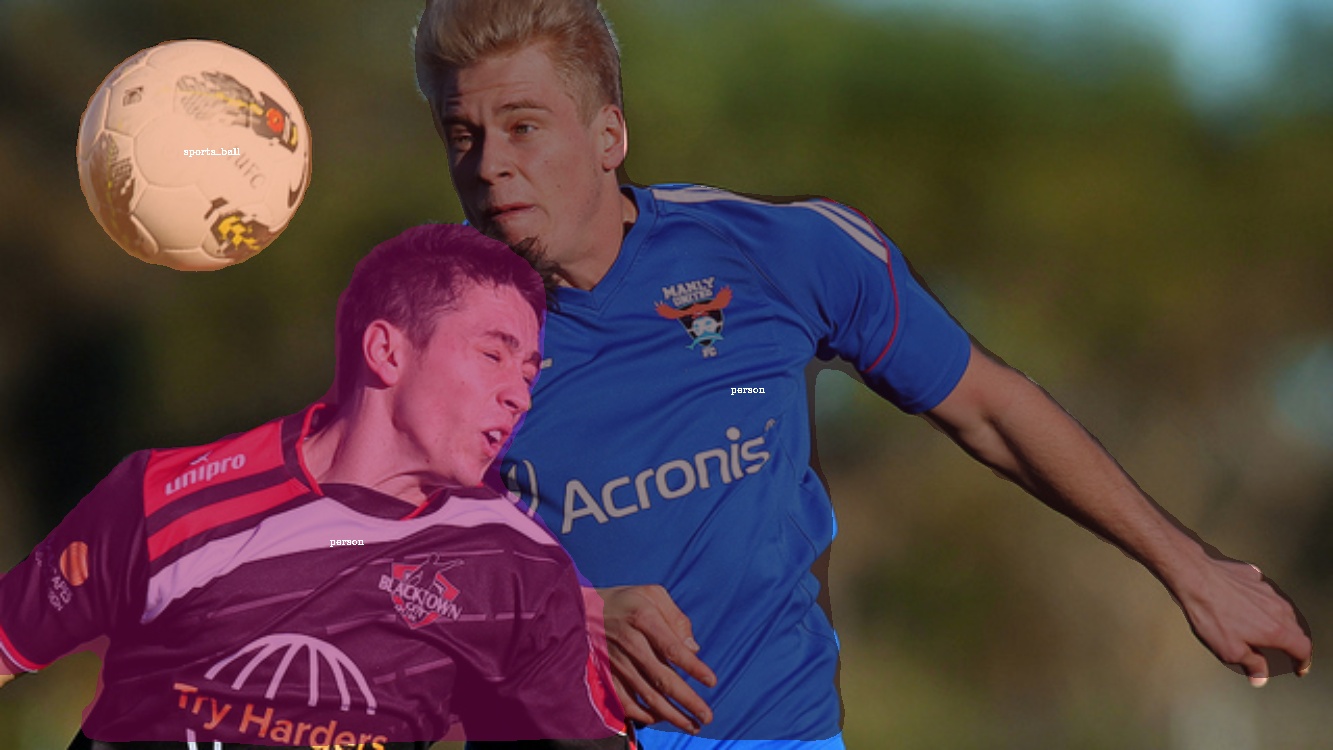}
\includegraphics[height=\visimgheightc]{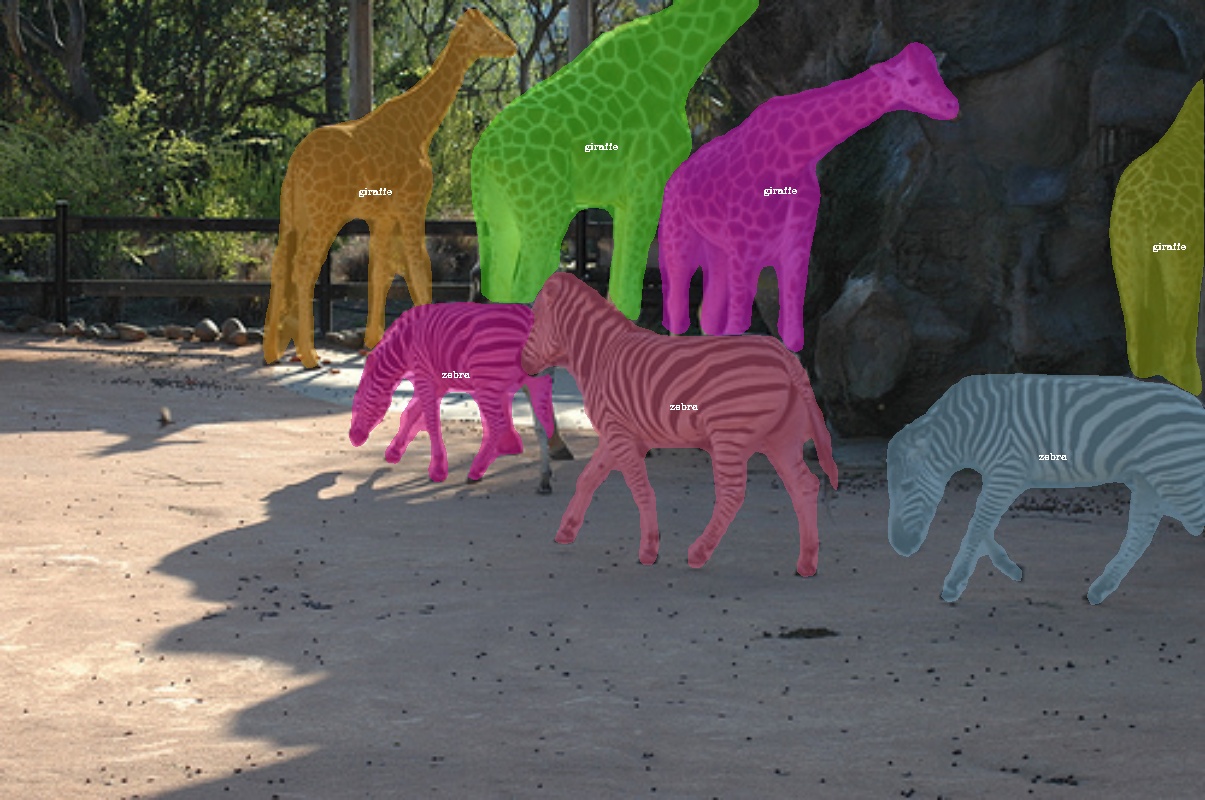}
\includegraphics[height=\visimgheightc]{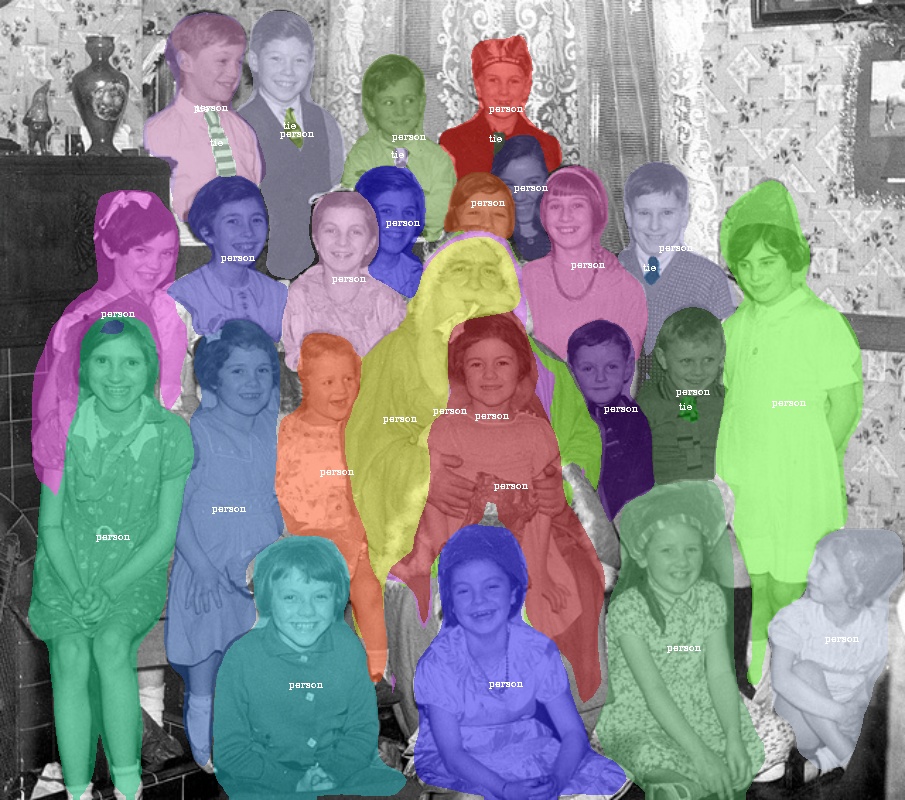}

\def\visimgheightd{2.72cm}
\includegraphics[height=\visimgheightd]{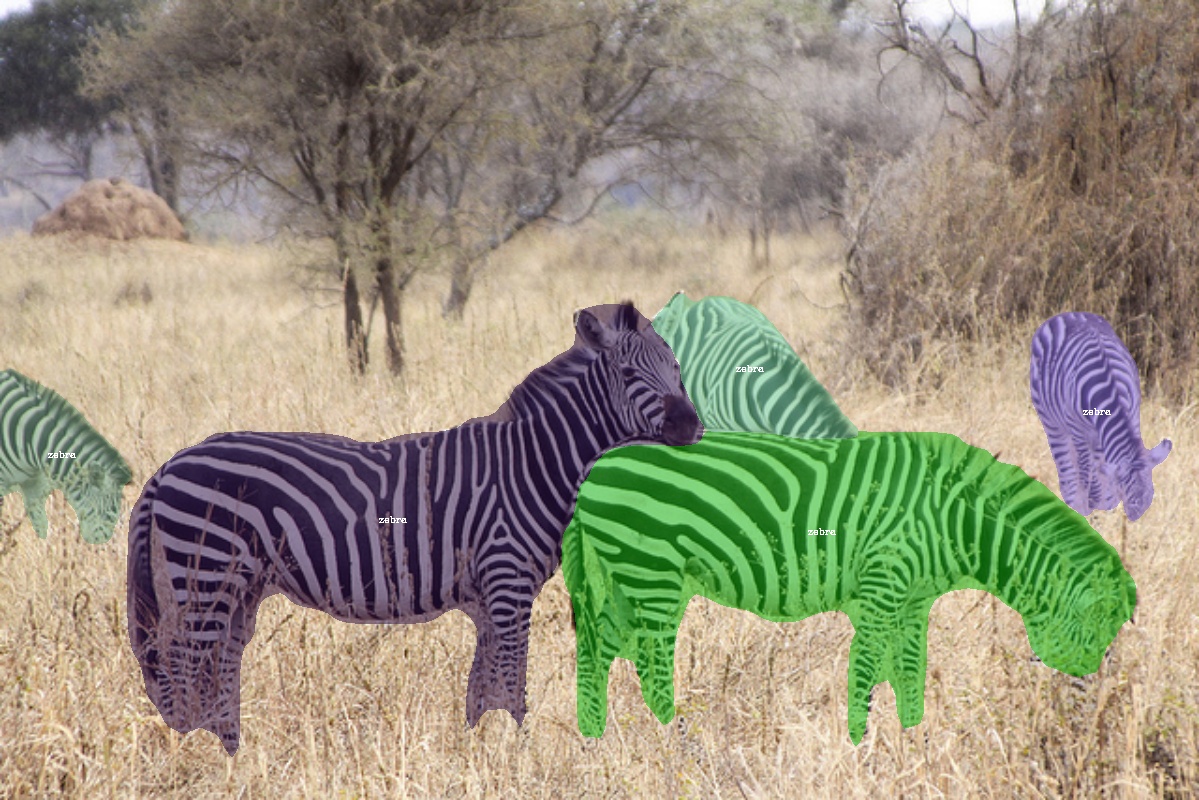}
\includegraphics[height=\visimgheightd]{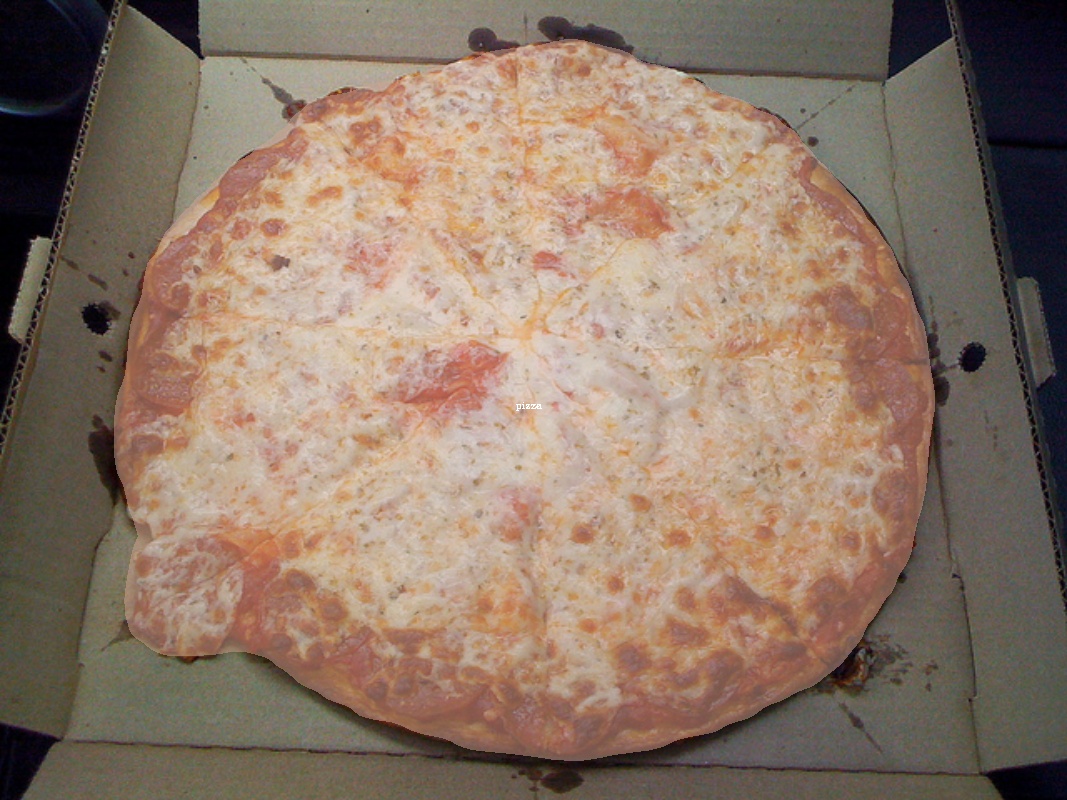}
\includegraphics[height=\visimgheightd]{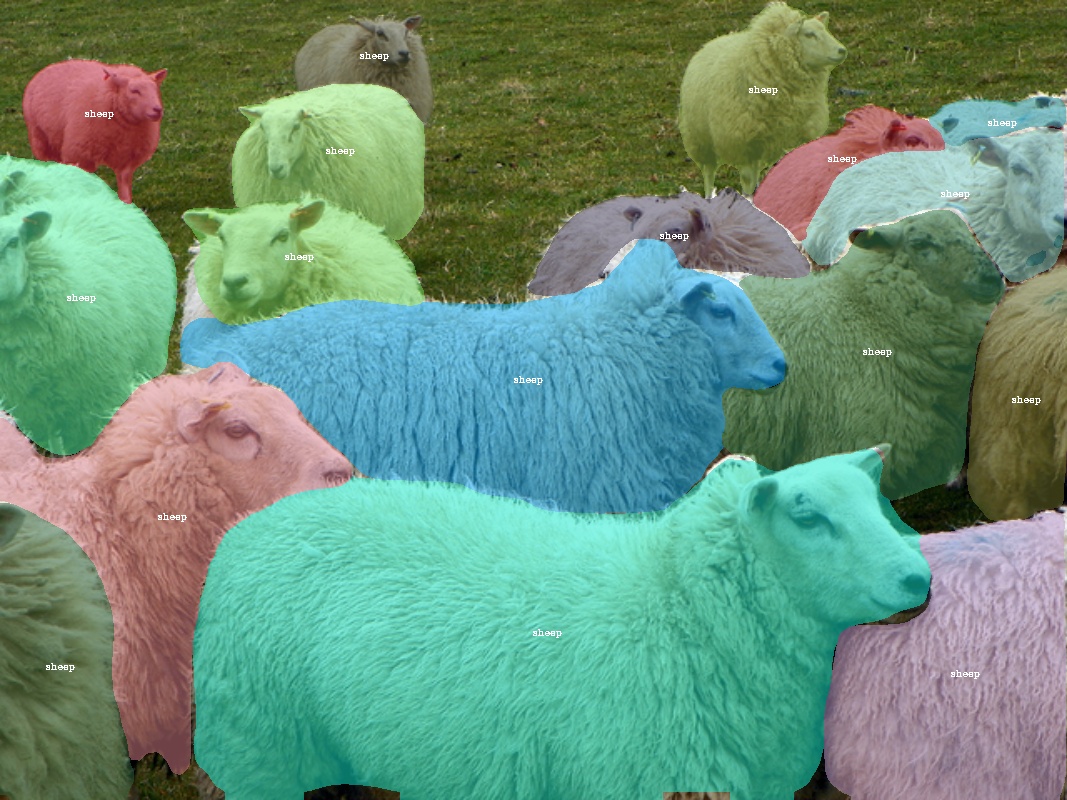}

\def\visimgheighte{2.56cm}
\includegraphics[height=\visimgheighte]{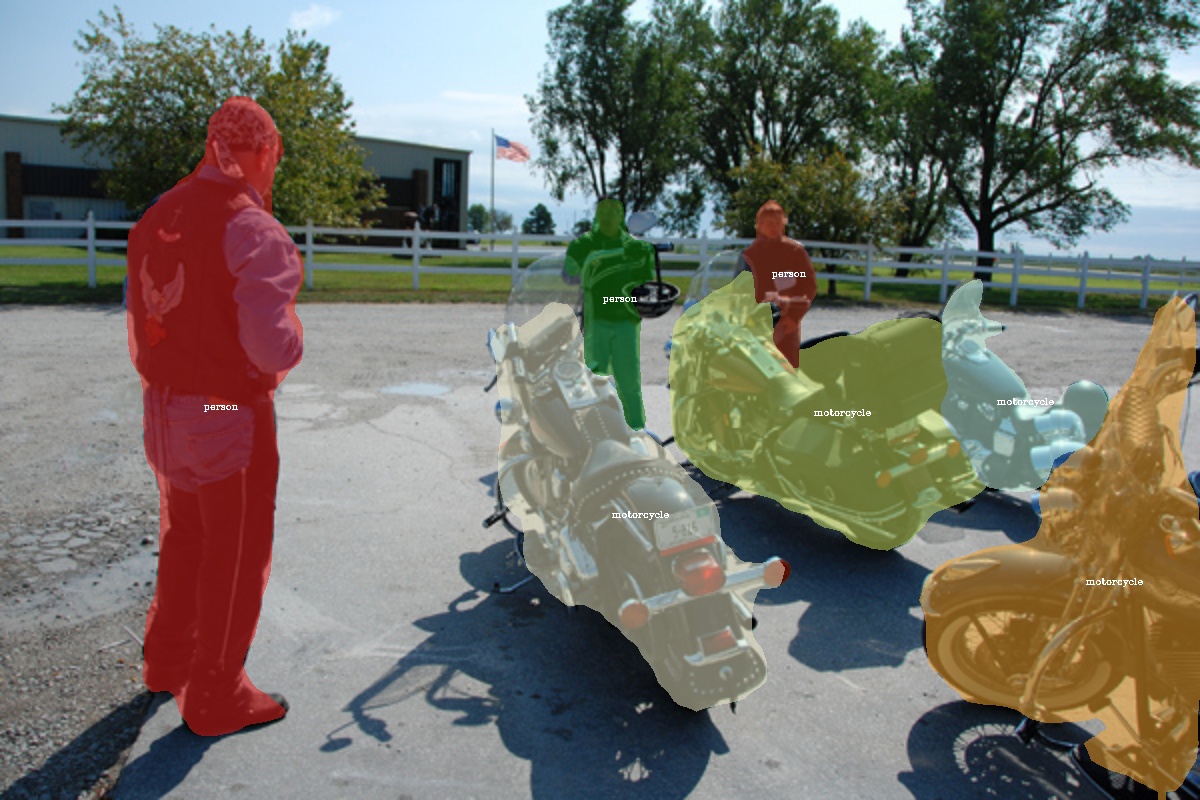}
\includegraphics[height=\visimgheighte]{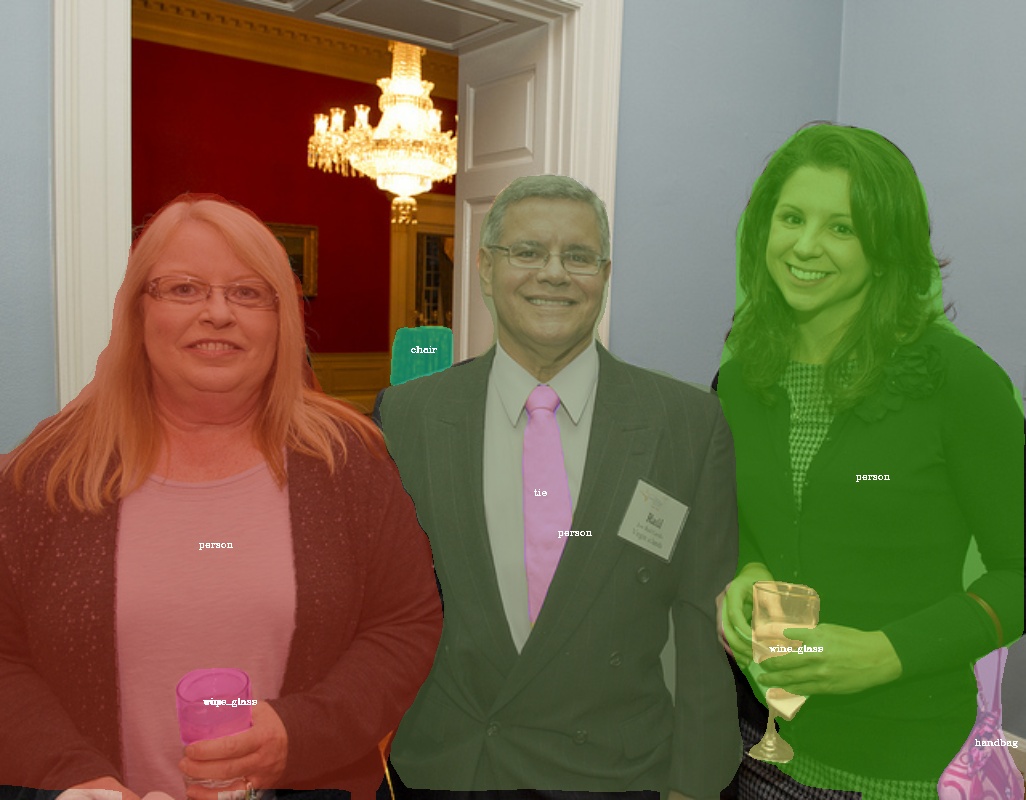}
\includegraphics[height=\visimgheighte]{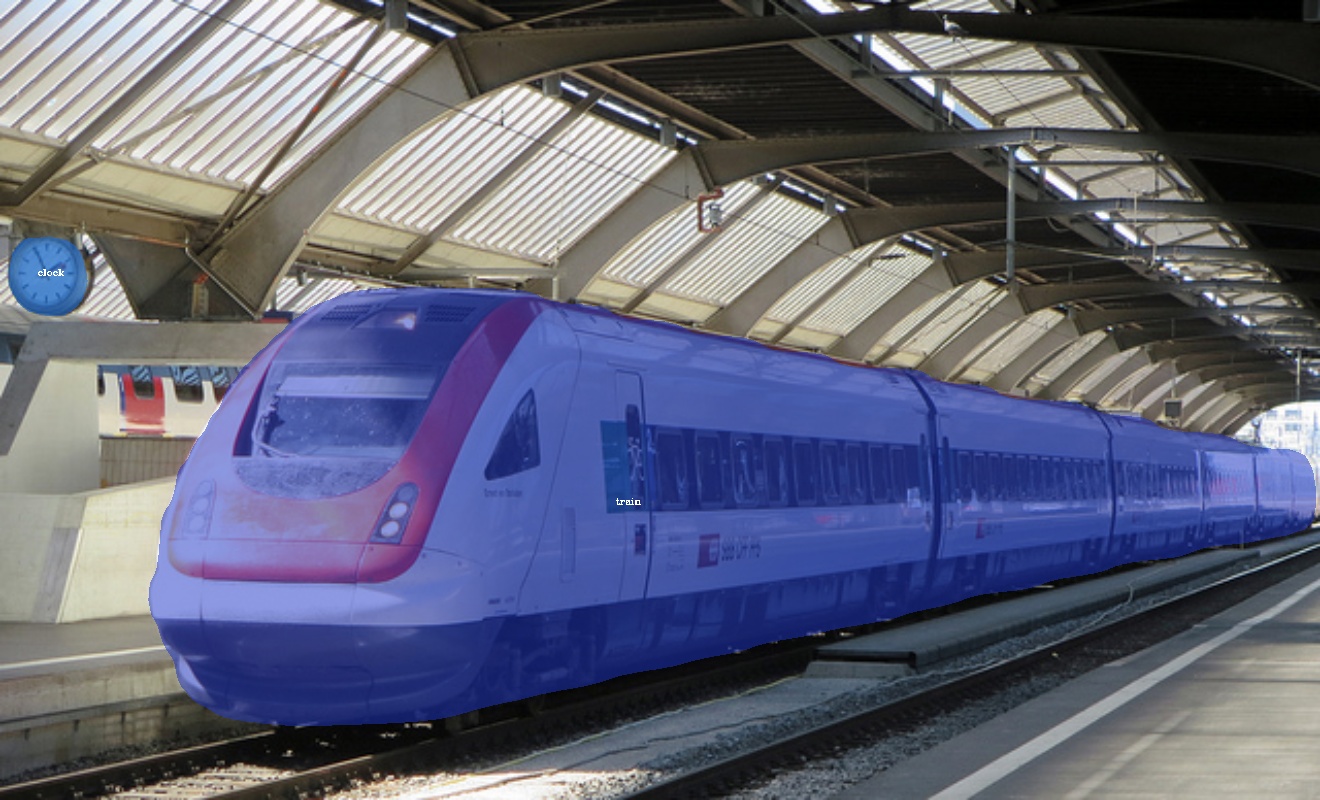}

\def\visimgheightf{2.65cm}
\includegraphics[height=\visimgheightf]{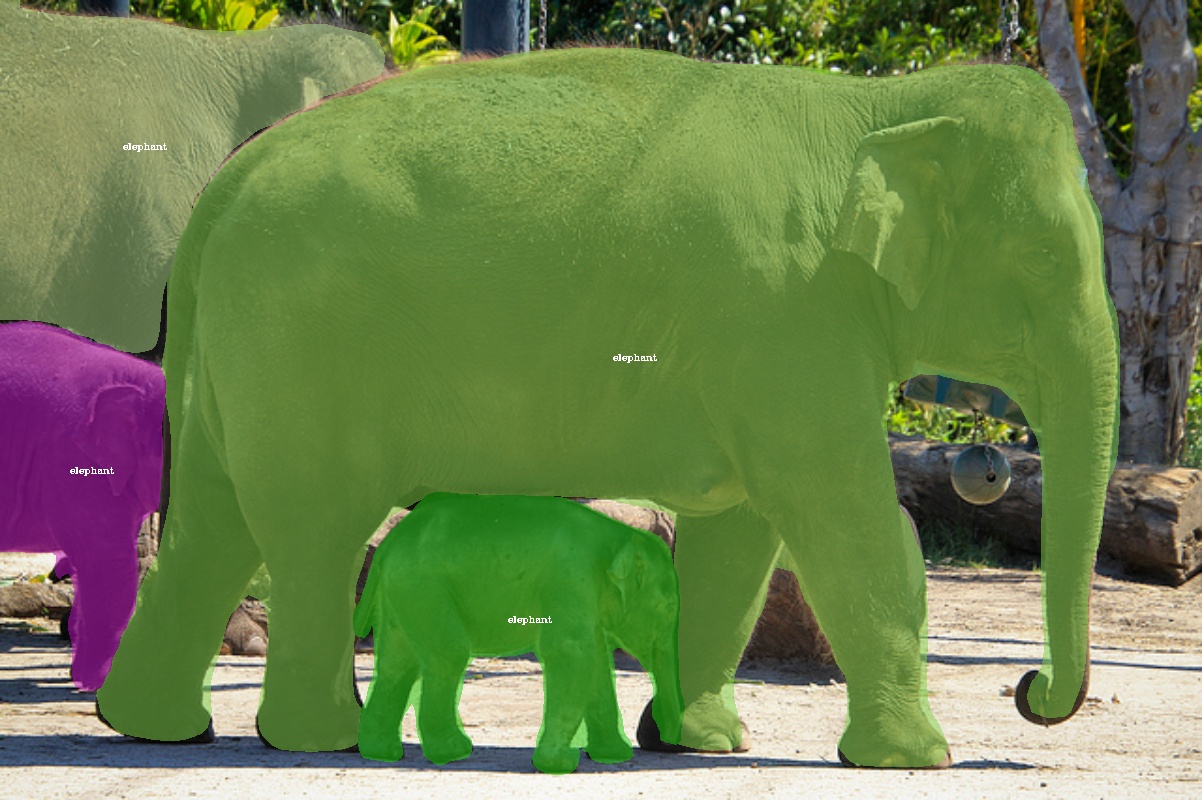}
\includegraphics[height=\visimgheightf]{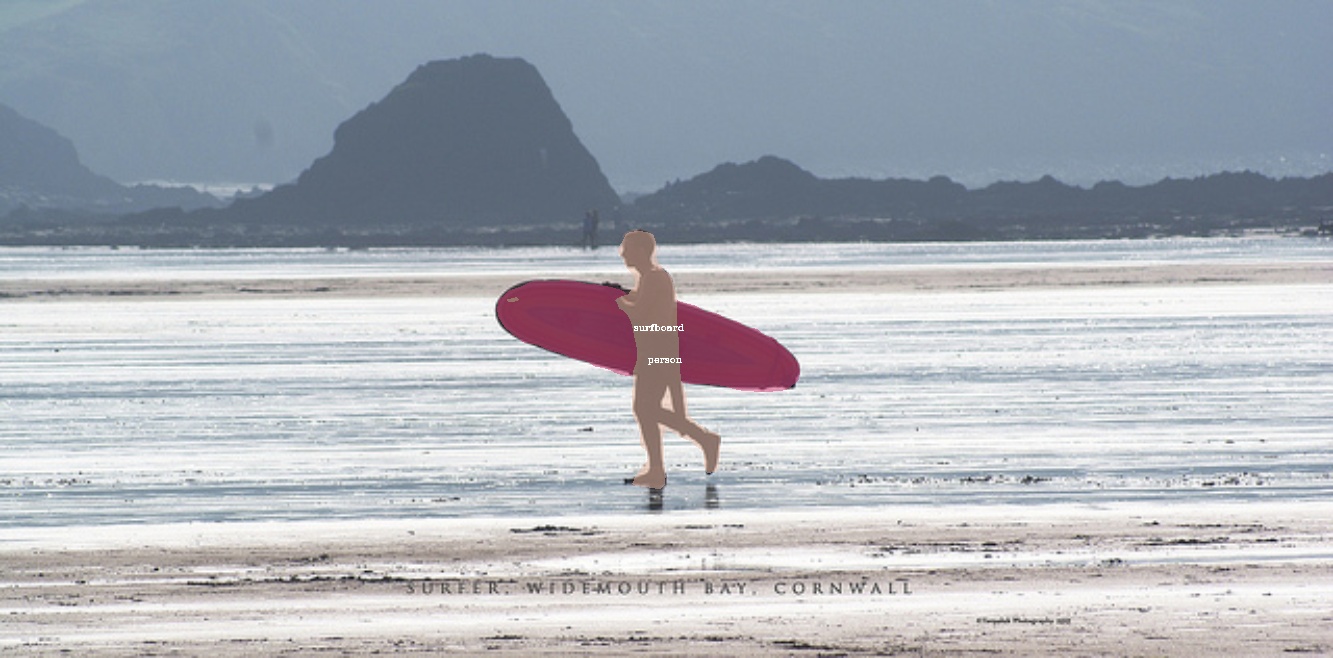}
\includegraphics[height=\visimgheightf]{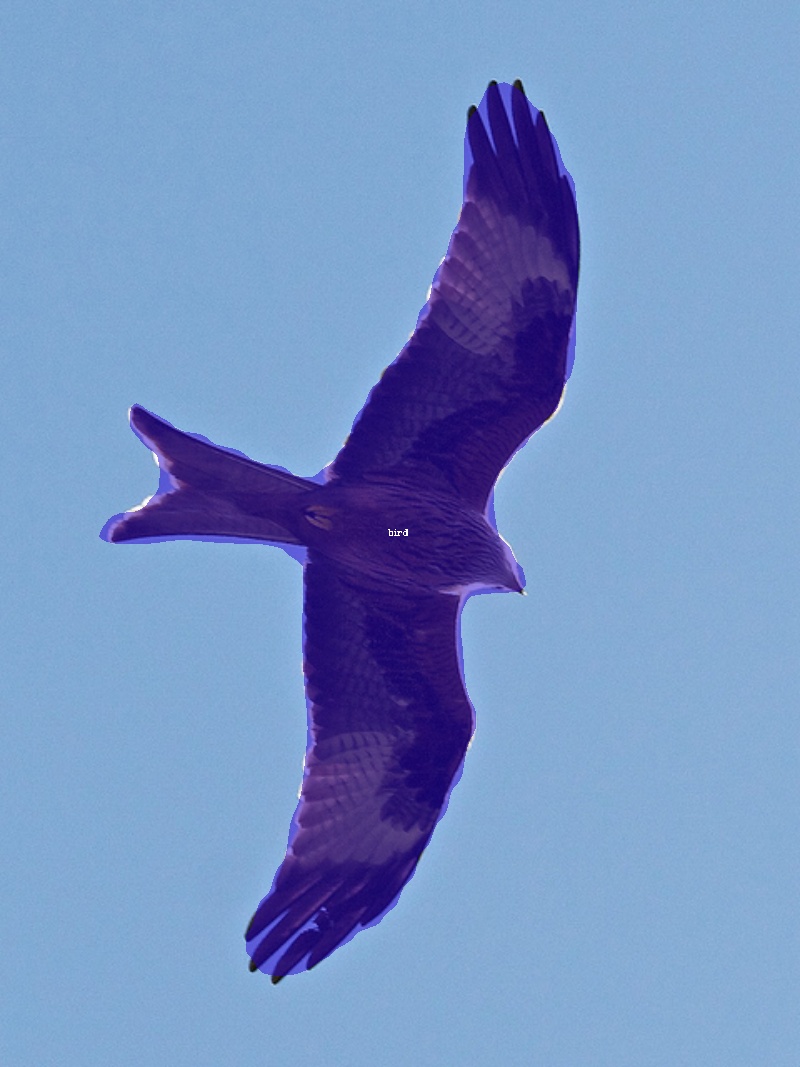}

} %
\caption{\textbf{Visualization of instance segmentation results}
using the Res-101-FPN backbone.
The model is trained on the COCO \texttt{train2017} dataset, achieving a mask AP of 37.8 on the COCO \texttt{test}-\texttt{dev}.
}
\label{fig:Vis}
\end{figure*}

\end{document}